\documentclass[11pt]{article}

\usepackage[final]{acl}

\usepackage{times}
\usepackage{latexsym}

\usepackage[T1]{fontenc}

\usepackage[utf8]{inputenc}

\usepackage{microtype}

\usepackage{inconsolata}

\usepackage{graphicx}

\usepackage{booktabs}
\usepackage{array}
\usepackage{amsmath}
\usepackage{textcomp}
\usepackage{pifont}
\usepackage{tabularx}
\usepackage{algorithm}
\usepackage{algpseudocode}
\usepackage{multirow}
\usepackage{paralist}
\usepackage{verbatim}
\usepackage{xcolor}
\usepackage{cuted}
\usepackage{tcolorbox}
\tcbuselibrary{breakable, skins}
\usepackage{icomma}

\tcbset{
  promptbox/.style={
    breakable,
    enhanced,
    colback=gray!10,
    colframe=gray!50,
    fonttitle=\bfseries,
    title={#1},
    left=6pt, right=6pt,
    top=4pt, bottom=4pt,
  }
}

\newcommand{\tacred}{FS-TACRED}
\newcommand{\fewrel}{FS-FewRel}
\newcommand{\qwensmall}{Qwen3-4B}
\newcommand{\qwenlarge}{Qwen3-14B}
\newcommand{\gemmasmall}{Gemma3-4B}
\newcommand{\gemmalarge}{Gemma3-12B}
\newcommand{\evoprompt}{EvoPrompt}
\newcommand{\rpo}{RPO}
\newcommand{\etgpo}{ETGPO}
\newcommand{\gradpo}{GradPO}
\newcommand{\gradpogen}{GradPO}
\newcommand{\gradpoprob}{GradPO-Prob}
\newcommand{\lpo}{LPO}
\newcommand{\greater}{GreaTer}

\usepackage{todonotes}

\title{Two-Stage Prompt Optimization for Few-Shot Relation Extraction: From Reasoning-Guided Search to Gradient-Guided Refinement}

\author{Aunabil Chakma, Mihai Surdeanu, and Eduardo Blanco\\
  University of Arizona, Tucson, AZ, USA \\
  \texttt{\{aunabilchakma, msurdeanu, eduardoblanco\}@arizona.edu} \\}

\begin{document}
\maketitle
\begin{abstract}
Automatic prompt optimization is still underexplored for episodic few-shot relation extraction with smaller language models.
We propose a two-stage framework that combines reasoning-based prompt optimization with gradient-based prompt optimization.
The first stage can use any reasoning-based optimizer to make broad prompt improvements in natural language.
The second stage applies our \gradpo{}, which uses loss and gradient signals to identify high-impact prompt spans and refine them with local edits.
Experiments on \tacred{} and \fewrel{} show that local refinement usually improves prompts found by the first stage, and \gradpo{} is the most consistent refiner.
Our framework achieves state-of-the-art performance on \tacred{} with \qwensmall{} and remains competitive on \fewrel{}.
\end{abstract}

\section{Introduction}
\label{sec:introduction}

Relation extraction (RE) identifies semantic relations between entity pairs in text.
It is a core task in information extraction and knowledge graph construction~\cite{IR_fundamental,KG_construction}.
Few-shot relation extraction (FSRE) is more challenging, as models must classify unseen relations from only a few support examples~\cite{fewrel}.
Large language models (LLMs) make FSRE attractive through in-context learning (ICL), where support examples are used directly at inference time without task-specific fine-tuning~\cite{GPT-RE,Unleash-ICL,struct-icl}.
Prior work improves ICL through better reasoning, relation-aware retrieval, and example selection, but the prompts themselves are still typically engineered manually rather than optimized automatically.

\begin{figure*}
    \centering
    \includegraphics[width=\textwidth]{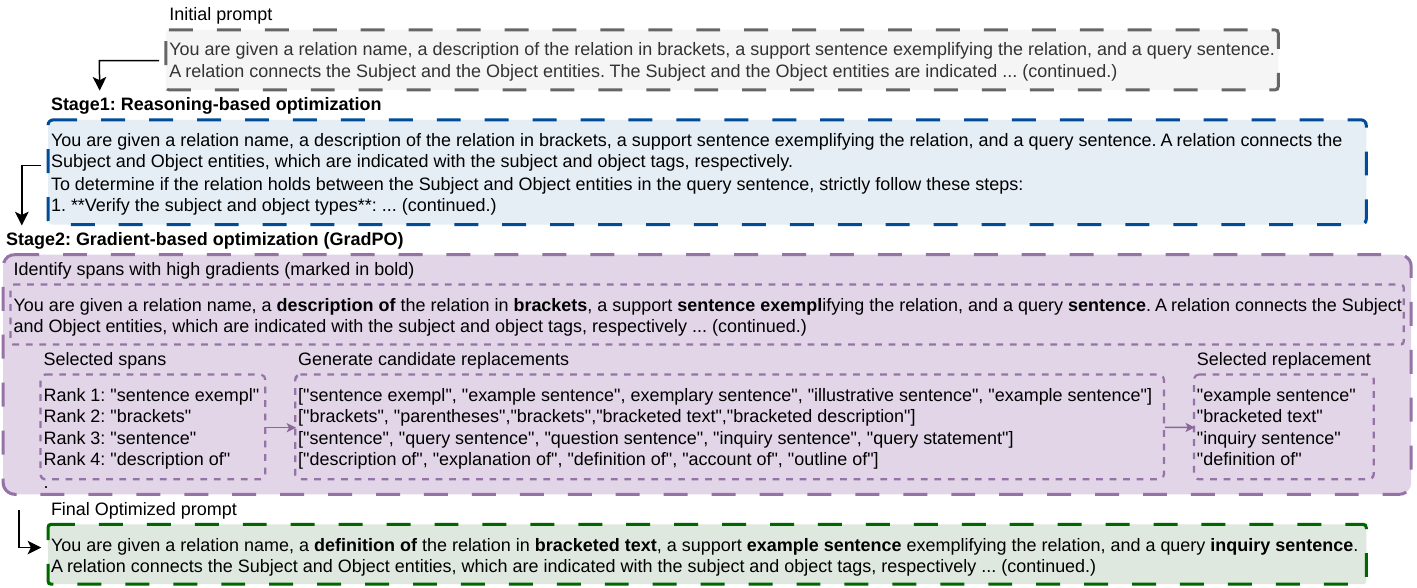}
    \caption{
    Overview of our two-stage prompt optimization framework for few-shot relation extraction, illustrated with a real optimization example using~\qwensmall{}.
    Starting from an initial prompt, the first stage performs Reasoning-Guided Prompt Optimization, 
    which can be instantiated with any automatic prompt optimization (APO) method, 
    to explore broad semantic improvements and produce an intermediate prompt.
    The second stage applies our novel Gradient-Guided Prompt Optimization, \gradpo, to the intermediate prompt, identifying high-impact prompt spans, 
    generating candidate replacements, and selecting the local edits that minimize the loss to obtain the final optimized prompt.
    The reasoning-guided stage efficiently improves the prompt over a few iterations, while the gradient-guided stage performs a targeted local refinement step.
    }
    \label{fig:framework}
\end{figure*}

Automatic Prompt optimization (APO) is an important field which can automatically optimize prompts~\cite{apo}.
Most prior RE work studies prompt tuning with soft inputs around pretrained encoders~\cite{related_work_1_knowprompt,CKPT}.
A recent exception optimizes natural-language prompts for relation triple extraction~\cite{related_work_4_automaticpromptoptimizationknowledge}, 
but that task is different from episodic FSRE, a more challenging task~\cite{fs-tacred}, 
where the prompt must work together with a small set of examples per relation in each episode.
Overall, APO remains underexplored for RE.

Outside RE, prompt optimization has mainly followed two directions: reasoning-based and gradient-based optimization.
Reasoning-based methods use errors or feedback to rewrite prompts in natural language~\cite{promptagent,textgrad}, enabling broad semantic and logical improvements.
Gradient-based methods, in contrast, do not directly learn from semantic or logical feedback; instead, they use loss and gradient signals to identify impactful tokens or spans for precise local edits~\cite{hard_prompt_made_easy,greater}.
This local refinement is important because LLM behavior can be highly sensitive to small token-level choices~\cite{llms_token_sensitivity}.
The two directions are therefore \emph{complementary}: reasoning-based optimization supports global prompt search from feedback, while gradient-based optimization provides fine-grained refinement.
However, most prior work studies them in isolation, leaving their combination underexplored in both RE and broader prompt optimization settings.

In this paper, we study automatic prompt optimization for episodic FSRE with smaller LLMs, 
formulated as a non-reasoning task (Section~\ref{sec:methodologies}).
We propose a two-stage framework that combines reasoning-based global search with gradient-guided local refinement.
To the best of our knowledge, this is the first work to systematically study this combination for natural-language prompt optimization in FSRE.
In the first stage, a reasoning-based optimizer uses feedback to make broad natural-language improvements.
In the second stage, we introduce \gradpo{}, a gradient-guided optimizer that \emph{identifies and edits multiple high-impact prompt spans} for local refinement.
For span replacement, we consider both LLM-generated replacements and token selection based on the target model's token probabilities.
Figure~\ref{fig:framework} illustrates the full pipeline with a real optimization example.
Our experiments show that the two stages are complementary: global search improves prompts quickly, while local refinement makes precise edits that further improve performance (Section~\ref{sec:results}).

Our primary contributions are as follows:
\begin{itemize}
    \item We study automatic prompt optimization for episodic FSRE and propose a two-stage framework 
    that combines reasoning-based prompt optimization for broad prompt improvement with gradient-based optimization for local refinement.
    \item We introduce \gradpo, a gradient-guided prompt optimizer that edits high-impact prompt spans 
    and can be plugged into any reasoning-based prompt optimizer as a second-stage refiner.
    Experiments on \tacred{} and \fewrel{} show that \gradpo{} improves prompts produced by different reasoning-based optimizers 
    and is the most consistent second-stage refiner.
    \item Our framework achieves state-of-the-art performance on \tacred{} with \qwensmall, improving F1 from 27.6 to 29.0, 
    while remaining competitive with prior work on \fewrel{}.
\end{itemize}
\section{Related Work}
\label{sec:related-works}

\paragraph{FSRE with ICL.}
Recent work shows that LLMs can perform relation extraction in the ICL setting through better prompting, reasoning, and demonstration design~\cite{COT-RE,Unleash-ICL,GPT-RE,C-ICL,struct-icl}.
These studies improve RE with chain-of-thought reasoning, relation-aware example retrieval, or contrastive examples, but they do not optimize the task prompt itself.
In this paper, we aim to study prompt optimizations for FSRE.

\paragraph{Prompt optimization for RE.}
Most prompt-based RE work focuses on soft prompts rather than rewriting natural-language prompts.
Representative methods use prompt tuning with soft prompts around pretrained encoders~\cite{related_work_1_knowprompt,CKPT}.
A notable exception is recent work on automatic prompt optimization for knowledge graph construction, which compares several natural-language prompt optimization methods~\cite{related_work_4_automaticpromptoptimizationknowledge}.
However, that work optimizes prompts for relation triple extraction, not episodic FSRE, which is a fundamentally different task.

\paragraph{Prompt optimization beyond RE.}
Outside RE, automatic prompt optimization has been studied more extensively.
Methods such as APO~\cite{apo}, PromptAgent~\cite{promptagent}, OPRO~\cite{OPRO}, and TextGrad~\cite{textgrad} 
iteratively refine prompts using scores, feedback, or search.
These methods often make broad prompt changes and can improve prompts in only a few iterations.
\lpo{}~\cite{lpo} also belongs to this broader line, but it narrows the search to a few local regions of the prompt instead of rewriting it globally.
We also employ a simple reasoning-based prompt optimization strategy tailored for RE for the first stage.

A separate line of work uses gradients for prompt optimization.
Some studies optimize prompts through continuous representations or gradient-aligned soft prompt updates 
rather than directly editing natural-language text~\cite{direct_PO,prompt_aligned_PT}.
Hard prompt optimization has also been studied with discrete gradient-based search over prompt tokens~\cite{hard_prompt_made_easy}.
Among recent methods, \greater{} is closest to our work 
because it uses gradients to optimize discrete prompt tokens while producing sensible natural-language prompts~\cite{greater}.
We extend this direction to local refinement over multiple selected prompt spans. 

\section{Two-stage Prompt Optimization} 
\label{sec:methodologies}

\subsection{Task Description}
We study few-shot relation extraction under the standard $N$-way $M$-shot episodic setting.
In our experiments, we focus on the one-shot setting, where $M=1$.
Each episode contains $N$ candidate relations, $M$ support examples for each relation, and one or more query instances.
The model must decide which relation, if any, is expressed in the query.
We cast episodic FSRE as $N$ binary relation classification problems.
For each candidate relation, the model is given the relation name, its description, one support sentence for that relation, and the query sentence, and it outputs either \texttt{yes} or \texttt{no}.
We repeat this process for all candidate relations in the episode and recover the final episodic prediction from the binary decisions.
If multiple relations are predicted as positive, we choose one at random.
If none of the candidate relations are predicted as positive, we output \texttt{no\_relation}\footnote{\texttt{no\_relation} means that none of the candidate relations is expressed between the subject and object entities in the query, not that the two entities have no relation at all.}.
Our goal is to optimize the prompt used for these binary decisions.

\subsection{Overview of Our Algorithm}
Our algorithm has two stages, as shown in Figure~\ref{fig:framework}.
In the first stage, we apply a reasoning-based prompt optimizer for global optimization.
We then pass the top-performing prompt, selected by validation score, to the second stage.
In the second stage, we apply gradient-based prompt optimization for local refinement.
We use three reasoning-based optimizers, \rpo{}, \evoprompt{}-DE, and \etgpo{}, in the first stage.
We use these optimized prompts as the starting point for \gradpo{} refinement, 
and compare \gradpo{} against two local refinement alternatives, \greater{} and \lpo{}.

\subsection{Global optimization via Reasoning-based prompt optimization}
\paragraph{RPO.}
Reasoning-based prompt optimization has become a common way to improve prompts in natural language space \cite{apo,OPRO,promptagent}.
We implement a relation extraction version of this idea and call it \rpo{}.
It maintains a small population of prompts and performs an iterative search.
At each iteration, a parent prompt is sampled from the current population according to a softmax distribution over validation scores.
We then sample training examples, run the parent prompt on them, and collect the model's predictions.
From these trials, we build a feedback set that contains representative successes and failures.
A separate optimizer model is then prompted to explain why the current prompt led to those correct or incorrect decisions.
These feedbacks are inserted into a mutation prompt, and the optimizer model generates a revised prompt.
The revised prompt is evaluated on the validation episodes.
We then add the new prompt to the population and prune weak prompts.
Because the mutation step is free to rewrite any part of the prompt, \rpo{} can make broad changes in only a few iterations.

\paragraph{EvoPrompt.}
We include a non-feedback-based prompt optimization method, \evoprompt{}~\cite{EVOPROMPT}.
It is an evolutionary approach that maintains a small population of prompts and generates new prompts by combining existing ones using a genetic algorithm.
In the original work, prompt fitness is evaluated directly on the validation set, which can cause overfitting in our task.
Instead, we evaluate each prompt's fitness on a training set and select the best prompt from the population for evaluation on the validation set in each iteration.
In our experiments, we only use the DE variant of \evoprompt{}, since it outperformed the other variant, GE, in the original work.

\paragraph{ETGPO.}
Finally, we study \etgpo{}~\cite{ETGPO}, which is also feedback-driven but does not run a search loop.
It first runs on large amount of samples to collect all the failures. In our implementation, these samples are drawn from the training set rather than the validation set. 
An optimizer model then summarizes the failures into an error taxonomy and selects the error categories that cover the most common problems.
Using this structured summary of errors, it generates a revised prompt in just using one step.

\subsection{Local Refinement via Gradient-Based Prompt Optimization}

\paragraph{\gradpo{}.}
After the first-stage reasoning-based optimizer finds a strong prompt, we further refine it with \gradpo{}, our proposed gradient-guided local prompt optimizer.
Unlike one-token editing methods (e.g., \greater),  \gradpo{} identifies and edits multiple high-impact spans in the prompt, allowing it to explore a richer local search space while preserving the overall prompt structure.
The overall procedure is shown in Algorithm~\ref{alg:local-gradient-refinement}.

Given the current prompt $P'$, \gradpo{} first samples a training subset $\mathcal{D}_{\mathrm{grad}}$ and computes gradients for each prompt token based on the loss on $\mathcal{D}_{\mathrm{grad}}$.
Rather than selecting a single token for editing, it groups nearby high-gradient tokens into editable spans.
Starting from the strongest gradient peaks, it expands each span to the left or right as long as neighboring tokens have sufficiently large gradient values under a threshold.
The resulting spans are ranked by their maximum gradient magnitude, and the top spans are selected for refinement.

\begin{algorithm}[t]
\caption{GradPO}
\label{alg:local-gradient-refinement}
\begin{algorithmic}[1]
\Require Initial prompt $P^{(0)}$, train set $\mathcal{D}_{\mathrm{train}}$, dev set $\mathcal{D}_{\mathrm{val}}$, iterations $Q$, beam size $M$, candidates per span $K$, mode $m \in \{\textsc{Gen}, \textsc{Prob}\}$
\Ensure Optimized prompt $P^{*}$

\For{$q = 1$ to $Q$}
    \State Sample $\mathcal{D}_{\mathrm{grad}} \subset \mathcal{D}_{\mathrm{train}}$
    \State Compute token gradients on $P^{(q-1)}$ using $\mathcal{D}_{\mathrm{grad}}$
    \State Select top spans $\mathcal{S}$ based on gradients

    \ForAll{$s=(i,j) \in \mathcal{S}$}
        \If{$m=\textsc{Gen}$}
            \State $C_s \gets \mathrm{LLM}(\operatorname{Mark}(P^{(q-1)}, s))$
        \ElsIf{$m=\textsc{Prob}$}
            \State $C_s \gets \operatorname{TopK}_{x_i}\big(p(x_i \mid P^{(q-1)}_{<i})\big)$
        \EndIf
    \EndFor

    \State $\mathcal{B} \gets \{P^{(q-1)}\}$

    \ForAll{$s \in \mathcal{S}$}
        \State $\mathcal{B}' \gets \{P_b[s \leftarrow c] \mid P_b \in \mathcal{B},\, c \in C_s\}$
        \State $\mathcal{B} \gets \operatorname{TopM}_{P \in \mathcal{B}'} \big(-\mathcal{L}(P,\mathcal{D}_{\mathrm{grad}})\big)$
    \EndFor

    \State $P^{(q)} \gets \arg\min_{P \in \mathcal{B}} \mathcal{L}(P,\mathcal{D}_{\mathrm{val}})$
\EndFor

\State $P^{*} \gets P^{(Q)}$
\State \Return $P^{*}$
\end{algorithmic}
\end{algorithm}

After selecting the spans, \gradpo{} generates candidate replacements for each span using LLM prompting.
Specifically, we mark the selected spans in the prompt and ask the target LLM to propose several natural-language replacements.
We further study another variant, \gradpoprob{}, where replacements are selected based on the target model's token probabilities.
For \gradpoprob{}, we use the left context of each selected span to sample high-probability first-token candidates, 
and then let the target model continue generation to complete the remaining tokens in the span.

To combine edits across multiple spans, \gradpo{} uses beam search.
The current prompt $P'$ is used as the initial beam.
For each selected span, \gradpo{} considers every candidate replacement for that span and applies it to each prompt in the current beam.
Because span replacement must preserve the natural-language structure of the prompt, we perform each replacement through prompting: 
given the current beam prompt and the candidate replacement, the optimizer LLM rewrites the prompt by substituting the selected span while keeping the rest of the prompt unchanged.
At each beam step, we ask the optimizer LLM to apply the candidate replacement for the current span on $P'$
together with the previously selected replacements for earlier spans, rather than editing the span in isolation.
The resulting prompts are then scored, and only the top beam-width candidate prompts are retained.

Following \greater{}, we score prompts using the following objective:
\begin{equation}
\mathcal{L}(p, \mathcal{D})=\mathcal{L}_{\mathrm{task}}(p, \mathcal{D})+\lambda \, \mathrm{PPL}(p),
\label{eq:local_refinement_loss}
\end{equation}
where \(\mathcal{L}_{\mathrm{task}}\) is the task loss on data $\mathcal{D}$, \(\mathrm{PPL}\) is a fluency penalty based on perplexity, and \(\lambda\) controls the penalty strength.
After all spans have been processed, the retained prompts are evaluated on the validation split, and the best prompt is selected as the refined prompt.

Finally, for both \gradpogen{} and \gradpoprob{}, we avoid using a separate optimizer model for candidate generation or span replacement, relying only on the target model itself.
Moreover, we disable reasoning during these prompting steps to keep the refinement process lightweight and non-reasoning-based.

\subsection{Alternative Baselines for Local Refinement}

None of the baselines discussed below was originally designed specifically as a local refiner.
We adapt them as strong baselines in our two-stage framework by applying them to prompts produced by the first-stage optimizer.

\paragraph{GreaTer.}
We compare \gradpo{} with \greater{}~\cite{greater}, a gradient-based prompt optimization method.
\greater{} was originally designed to iteratively optimize the full prompt by editing one token at a time, rather than to refine selected spans after a global optimizer.
For a selected token position, \greater{} proposes candidate replacement tokens based on language-model probabilities and 
scores each resulting prompt using the objective in Equation~\ref{eq:local_refinement_loss} on a sampled training subset $\mathcal{D}_{\mathrm{grad}}$.
It then keeps the top token replacements with the lowest combined loss.
Similar to \gradpogen{}, \greater{} does not use a separate optimizer model.

The original \greater{} edits prompt tokens sequentially in prompt order.
We also test a top-gradient variant that recomputes gradients at each step and edits the token with the largest gradient magnitude.
We refer to this variant as \greater{}-TG.
In both variants, the search space is highly local because each iteration changes only one token.
This makes \greater{} a useful baseline for precise but narrow gradient-based refinement.

\paragraph{LPO.}
We also compare against \lpo{}~\cite{lpo} as a gradient-free local refinement baseline.
At each iteration, \lpo{} first collects errors on sampled training examples and uses prompting to identify multiple short spans for editing.
It then generates multiple revised prompts by rewriting only those marked spans through prompting while keeping the rest of the prompt fixed.
The revised prompts are evaluated on the validation set, and the best-performing prompt among the generated candidates is selected for the next iteration.

Additional details, including the hyperparameters, meta-prompts, and 
prompting templates used for all optimization methods, 
are provided in Appendix Sections~\ref{sec:additional_details_opt}, \ref{subsec:meta-prompts}, \ref{sec:prompt_template}, and~\ref{subsec:optimized-prompts}. 

\section{Experiment Setup}
\label{sec:experiment-setup}

\paragraph{LLMs.}
We use \qwensmall{} and \gemmasmall{} as the target models in all main experiments.
For the non-gradient based methods, we use the corresponding larger models as optimizers (\qwenlarge{} and \gemmalarge).
However, for \rpo{}, we use the \qwensmall{} as optimizer since the optimized prompts were better than the optimized prompts by \qwenlarge{} in validation set.

\paragraph{Datasets.}
We evaluate on \tacred{} and \fewrel{}.
\tacred{} is the realistic few-shot benchmark introduced by~\citet{fs-tacred}.
\fewrel{} is the episodic benchmark created by~\citet{struct-icl} from the FewRel dataset~\cite{fewrel}, for direct comparison with their work.
Both datasets are in $5$-way $1$-shot episodic form with five sets of $10,000$ episodes.
The setting is challenging because the model sees only one support example per candidate relation.
It is especially hard because most of the queries do not match any candidate relation and must be mapped to \texttt{no\_relation}
\footnote{$97\%$ and $95\%$ of episodes of \tacred{} and \fewrel{} have no\_relation as gold truth labels.}.
Additional details about the train and validation split construction are provided in Appendix~\ref{sec:dataset_split_details}.

\paragraph{Hyperparameter Optimization.}
Hyperparameters associated with different 2nd-stage optimizers are tuned only on \tacred{} using prompts produced by \rpo{} for each model, and the same settings are reused for the remaining experiments.
Additional details and hyperparameters are provided in Appendix~\ref{sec:additional_details_opt}.

\paragraph{Metrics and Evaluation.}
After optimization, the top prompt from each method is selected using validation performance for test-set inference.
We report the averages of precision, recall, and F1 across five runs, and use mean F1 as the primary selection and comparison metric.
In the final tables, we report the mean, standard deviation, and statistical significance results under each method's evaluation protocol. 

\section{Results}
\label{sec:results}
Tables~\ref{tab:results_main_tacred}, \ref{tab:results_main_fewrel}, and \ref{tab:results_etgpo} report the main results with our two-stage prompt optimization approach on \tacred{} and \fewrel{}, 
evaluating whether local refinement consistently improves prompts produced by different reasoning-based optimizers.

\begin{table*}[t]
  \centering
  \small
  \setlength{\tabcolsep}{.061in}
\begin{tabular}{l r@{+}l
                r@{ \scriptsize$\pm$ }l r@{ \scriptsize$\pm$ }l l@{ \scriptsize$\pm$ }l
                r@{ \scriptsize$\pm$ }l r@{ \scriptsize$\pm$ }l l@{ \scriptsize$\pm$ }l}
\toprule
  &
  \multicolumn{2}{c}{\#itr.} &
  \multicolumn{6}{c}{Qwen3-4B} &
  \multicolumn{6}{c}{Gemma3-4B} \\ \cmidrule(lr){4-9} \cmidrule(lr){10-15}
& R & G &
  \multicolumn{2}{c}{P} & \multicolumn{2}{c}{R} & \multicolumn{2}{c}{F1} &
  \multicolumn{2}{c}{P} & \multicolumn{2}{c}{R} & \multicolumn{2}{c}{F1} \\ \midrule

Initial Prompt     & \multicolumn{2}{c}{} & 17.6 & 0.44 & 38.2 & 1.38 & 24.0 & 0.59 & 10.4 & 0.56 & 12.5 & 0.65 & 11.3 & 0.53 \\ 
\midrule

RPO
& 5 & 0 & 30.8 & 1.12 & 23.3 & 1.34 & 26.5$^{\dagger}$ & 1.19 & 15.8 & 1.12 & 12.4 & 0.83 & 13.9$^{\dagger}$ & 0.94 \\
~~~~~+LPO
& 5 & 1 & 30.4 & 1.41 & 21.0 & 0.90 & 24.8$^{\dagger}$ & 0.98 & 15.1 & 0.63 & 15.0 & 0.49 & \textbf{15.0} & 0.54 \\
~~~~~+GreaTer
& 5 & 1 & 32.1 & 1.44 & 21.8 & 1.58 & 25.9$^{\dagger}$ & 1.43 & 14.6 & 1.11 & 13.5 & 0.96 & 14.0$^{\dagger}$ & 1.01 \\
~~~~~+GreaTer-TG
& 5 & 1 & 27.9 & 1.58 & 26.5 & 1.69 & 27.2$^{\dagger}$ & 1.53 &  \multicolumn{2}{c}{-} & \multicolumn{2}{c}{-} & \multicolumn{2}{c}{-}\\
~~~~~+GradPO-Gen
& 5 & 1 & 29.5 & 1.25 & 28.5 & 1.66 & \textbf{29.0} & 1.37 & 15.9 & 1.25 & 12.4 & 0.95 & 13.9$^{\dagger}$ & 1.06 \\
~~~~~+GradPO-Prob
& 5 & 1 & 29.9 & 1.04 & 25.3 & 1.05 & 27.4$^{\dagger}$ & 0.94 &  \multicolumn{2}{c}{-} & \multicolumn{2}{c}{-} & \multicolumn{2}{c}{-}\\
\\
RPO
& 10 & 0 & 35.8 & 1.19 & 23.4 & 1.26 & 28.3$^{\dagger}$ & 1.13 & 09.8 & 0.33 & 14.4 & 0.72 & 11.7$^{\dagger}$ & 0.45 \\
~~~~~+LPO
& 10 & 1 & 35.3 & 1.71 & 29.1 & 1.72 & \textbf{31.9} & 1.65 &  \multicolumn{2}{c}{-} & \multicolumn{2}{c}{-} & \multicolumn{2}{c}{-}\\
~~~~~+GreaTer
& 10 & 1 & 35.8 & 1.18 & 24.1 & 1.57 & 28.8$^{\dagger}$ & 1.38 & 09.6 & 0.31 & 12.5 & 0.31 & 10.9$^{\dagger}$ & 0.27 \\
~~~~~+GreaTer-TG
& 10 & 1 & 37.7 & 1.76 & 21.0 & 1.15 & 26.9$^{\dagger}$ & 1.21 & 09.8 & 0.27 & 15.6 & 0.79 & \textbf{12.0} & 0.43 \\
~~~~~+GradPO-Gen
& 10 & 1 & 35.3 & 1.36 & 24.5 & 1.53 & 28.9$^{\dagger}$ & 1.46 & 10.2 & 0.20 & 14.3 & 0.31 & 11.9 & 0.11 \\
~~~~~+GradPO-Prob
& 10 & 1 & 37.9 & 1.24 & 22.7 & 0.73 & 28.4$^{\dagger}$ & 0.68 &  \multicolumn{2}{c}{-} & \multicolumn{2}{c}{-} & \multicolumn{2}{c}{-}\\

\midrule

EvoPrompt-DE
& 5 & 0 & 19.4 & 0.80 & 33.7 & 2.24 & 24.6 & 1.21 & 11.7 & 0.55 & 19.3 & 0.55 & \textbf{14.5} & 0.53 \\
~~~~~+LPO
& 5 & 1 &  \multicolumn{2}{c}{-} & \multicolumn{2}{c}{-} & \multicolumn{2}{c}{-} & 13.7 & 0.70 & 12.7 & 0.20 & 13.1$^{\dagger}$ & 0.27 \\
~~~~~+GreaTer
& 5 & 1 & 20.0 & 0.72 & 30.5 & 1.81 & 24.2$^{\dagger}$ & 1.04 &  \multicolumn{2}{c}{-} & \multicolumn{2}{c}{-} & \multicolumn{2}{c}{-}\\
~~~~~+GreaTer-TG
& 5 & 1 & 20.3 & 1.06 & 29.6 & 1.90 & 24.1$^{\dagger}$ & 1.31 & 11.4 & 0.71 & 17.9 & 0.81 & 13.9$^{\dagger}$ & 0.72 \\
~~~~~+GradPO-Gen
& 5 & 1 & 20.7 & 0.83 & 31.1 & 1.49 & \textbf{24.8} & 0.99 &  \multicolumn{2}{c}{-} & \multicolumn{2}{c}{-} & \multicolumn{2}{c}{-}\\
~~~~~+GradPO-Prob
& 5 & 1 & 19.9 & 0.79 & 33.0 & 1.86 & \textbf{24.8} & 1.08 & 11.7 & 0.68 & 19.1 & 0.80 & \textbf{14.5} & 0.72 \\
\\
EvoPrompt-DE
& 10 & 0 & 21.6 & 0.53 & 28.4 & 1.77 & 24.5 & 0.98 & 12.5 & 0.54 & 08.5 & 0.48 & 10.1$^{\dagger}$ & 0.42 \\
~~~~~+LPO
& 10 & 1 & 21.0 & 0.95 & 29.2 & 1.93 & 24.4 & 1.26 & 11.4 & 0.74 & 09.6 & 0.57 & 10.4 & 0.59 \\
~~~~~+GreaTer
& 10 & 1 & 22.1 & 0.38 & 28.2 & 1.44 & \textbf{24.7} & 0.78 & 12.3 & 0.70 & 08.7 & 0.39 & 10.2$^{*}$ & 0.39 \\
~~~~~+GreaTer-TG
& 10 & 1 &  \multicolumn{2}{c}{-} & \multicolumn{2}{c}{-} & \multicolumn{2}{c}{-} & 12.6 & 0.59 & 09.0 & 0.49 & \textbf{10.5} & 0.39 \\
~~~~~+GradPO-Gen
& 10 & 1 & 18.5 & 0.54 & 32.1 & 1.56 & 23.4$^{\dagger}$ & 0.74 & 12.2 & 0.77 & 08.9 & 0.44 & 10.3 & 0.50 \\
~~~~~+GradPO-Prob
& 10 & 1 & \multicolumn{2}{c}{-} & \multicolumn{2}{c}{-} & \multicolumn{2}{c}{-} & 12.4 & 0.58 & 08.3 & 0.51 & 09.9$^{\dagger}$ & 0.46 \\

\bottomrule

\end{tabular}
  \caption{Results on \tacred{} with \rpo{} and \evoprompt{}-DE as first-stage optimizers, and \lpo{}, \greater{}, and \gradpo{} as second-stage refiners.
  R and G denote reasoning-based and local refinement iterations, respectively.
  We report precision, recall, and F1 as mean \(\pm\) standard deviation, and bold marks the best F1 within each first-stage block.
  $^{*}$ and $^{\dagger}$ mark scores that are statistically significantly better within the same first-stage block at $p<0.05$ and $p<0.01$, respectively.
  Missing values indicate that the refined prompt did not improve over the stage-1 prompt on the dev set, so we omit test evaluation.
  }  
  \label{tab:results_main_tacred}
\end{table*}

\paragraph{Gradient-based refinement improves further after reasoning-based optimization.}
Tables~\ref{tab:results_main_tacred} and~\ref{tab:results_main_fewrel}, together with Table~\ref{tab:results_etgpo}, show that a second-stage local refinement is usually helpful after a first-stage reasoning-based optimizer.
Here, the first-stage optimizer is one of \rpo{}, \evoprompt{}, or \etgpo{}, and the second-stage alternatives are \gradpogen{}, \lpo{}, and \greater{}.
Across the 19 cases\footnote{There are 20 possible stage-2 cases we experimented in total, but one \evoprompt{} prompt after 5 iterations did not improve over the initial prompt for \qwensmall{} on \fewrel{}, so we do not refine it further}, 
17 are improved by at least one second-stage refinement method.
This shows that prompts found by the first-stage search often still contain locally suboptimal wording choices that can be improved afterward.
Among the second-stage refiners, our \gradpogen{} is the most consistent and improves 14 of the 19 prompts.
After that, \lpo{} and the original \greater{} variant each improve 8 prompts.

\paragraph{\gradpo{} outperforms other alternatives for refinement.} 
The same tables show that \gradpogen{} gives larger gains than the other second-stage alternatives in most cases.
Among the 17 prompts that are improved by at least one refiner, \gradpogen{} gives the best improvement 8 times.
In comparison, \lpo{} gives the best improvement 4 times, and the original \greater{} gives the best improvement 3 times.
Across different first-stage optimizers, \gradpogen{} is stronger after \rpo{} and \evoprompt{}, while \greater{} is slightly better after \etgpo{}.
Even with that exception, \gradpogen{} is the overall strongest second-stage refinement method in our experiments.

\paragraph{\gradpo{} is stable for most cases.}  
Table~\ref{tab:gradpo_stability} shows that first-stage reasoning-based optimization remains unstable across iterations, 
although the later iterations still give slight average improvements.
In each range, the average F1 is still much lower than the best F1, which means many prompts in that range perform substantially worse than the best one.
In contrast, a single \gradpogen{} step after iteration 10 improves the best prompt from the first 10 iterations in almost all cases, with only one degradation in the table.
This suggests that \gradpo{} can often provide a stable local improvement after the first-stage search has already found a strong prompt.
This benefits from searching over a larger local edit space before selecting the best refinement.
Overall, these results support the role of \gradpo{} as a complementary second-stage refiner rather than a replacement for reasoning-based optimization.

\paragraph{Our method improves over previous works on \tacred{} with \qwensmall.} 
Table~\ref{tab:results_tacred_revious_works} shows that our prompt optimization approach sets a new state of the art on \tacred{} with \qwensmall.
The best previous result is 27.6 F1, while our best setting in Table~\ref{tab:results_main_tacred} is higher with a score of 29.0.
On \fewrel{}, our approach also remains competitive: when combined with \evoprompt{}, it reaches 37.5 F1 in Table~\ref{tab:results_main_fewrel}, compared with 38.4 by~\citet{struct-icl}.

\begin{table*}[t]
  \centering
  \small
  \setlength{\tabcolsep}{.061in}
\begin{tabular}{l r@{+}l
                r@{ \scriptsize$\pm$ }l r@{ \scriptsize$\pm$ }l l@{ \scriptsize$\pm$ }l
                r@{ \scriptsize$\pm$ }l r@{ \scriptsize$\pm$ }l l@{ \scriptsize$\pm$ }l}
\toprule
 &
  \multicolumn{2}{c}{\#itr.} &
  \multicolumn{6}{c}{Qwen3-4B} &
  \multicolumn{6}{c}{Gemma3-4B} \\ \cmidrule(lr){4-9} \cmidrule(lr){10-15}
& R & G &
  \multicolumn{2}{c}{P} & \multicolumn{2}{c}{R} & \multicolumn{2}{c}{F1} &
  \multicolumn{2}{c}{P} & \multicolumn{2}{c}{R} & \multicolumn{2}{c}{F1} \\ \midrule

Initial Prompt             & \multicolumn{2}{c}{} & 29.9 & 0.72 & 44.1 & 0.76 & 35.6 & 0.66 & 39.9 & 2.13 & 18.6 & 1.12 & 25.3 & 1.27 \\ \midrule

RPO
& 5 & 0 & 42.4 & 0.97 & 29.5 & 0.55 & 34.8$^{\dagger}$ & 0.60 & 35.1 & 1.06 & 26.7 & 1.27 & 30.3$^{\dagger}$ & 1.14 \\
~~~~~+LPO
& 5 & 1 & 43.2 & 0.91 & 29.9 & 0.79 & 35.3$^{*}$ & 0.82 & 32.8 & 1.11 & 43.8 & 2.03 & \textbf{37.5} & 1.30 \\
~~~~~+GreaTer
& 5 & 1 & 42.2 & 1.32 & 29.1 & 0.88 & 34.5$^{\dagger}$ & 0.99 & 35.0 & 1.13 & 26.5 & 1.35 & 30.2$^{\dagger}$ & 1.23 \\
~~~~~+GreaTer-TG
& 5 & 1 & \multicolumn{2}{c}{-} & \multicolumn{2}{c}{-} & \multicolumn{2}{c}{-} & \multicolumn{2}{c}{-} & \multicolumn{2}{c}{-} & \multicolumn{2}{c}{-} \\
~~~~~+GradPO-Gen
& 5 & 1 & 42.9 & 1.27 & 30.7 & 0.64 & \textbf{35.8} & 0.80 & 30.2 & 0.85 & 35.7 & 1.70 & 32.7$^{\dagger}$ & 1.12 \\
~~~~~+GradPO-Prob
& 5 & 1 & \multicolumn{2}{c}{-} & \multicolumn{2}{c}{-} & \multicolumn{2}{c}{-} & \multicolumn{2}{c}{-} & \multicolumn{2}{c}{-} & \multicolumn{2}{c}{-} \\
\\
RPO
& 10 & 0 & 33.5 & 0.98 & 39.6 & 0.66 & 36.3$^{\dagger}$ & 0.74 & 34.8 & 1.39 & 29.9 & 1.42 & 32.2$^{\dagger}$ & 1.30 \\
~~~~~+LPO
& 10 & 1 & \multicolumn{2}{c}{-} & \multicolumn{2}{c}{-} & \multicolumn{2}{c}{-} & \multicolumn{2}{c}{-} & \multicolumn{2}{c}{-} & \multicolumn{2}{c}{-} \\
~~~~~+GreaTer
& 10 & 1 & \multicolumn{2}{c}{-} & \multicolumn{2}{c}{-} & \multicolumn{2}{c}{-} & \multicolumn{2}{c}{-} & \multicolumn{2}{c}{-} & \multicolumn{2}{c}{-} \\
~~~~~+GreaTer-TG
& 10 & 1 & 32.8 & 0.93 & 40.9 & 0.70 & 36.4$^{\dagger}$ & 0.74 & 34.6 & 1.20 & 31.1 & 1.47 & 32.7$^{*}$ & 1.25 \\
~~~~~+GradPO-Gen
& 10 & 1 & 34.2 & 0.94 & 40.2 & 0.94 & \textbf{36.9} & 0.82 & 32.5 & 1.28 & 34.1 & 1.77 & \textbf{33.3} & 1.40 \\
~~~~~+GradPO-Prob
& 10 & 1 & 33.1 & 1.03 & 40.8 & 0.48 & 36.6$^{*}$ & 0.76 & 35.2 & 1.19 & 29.0 & 1.54 & 31.8$^{\dagger}$ & 1.28 \\

\midrule

EvoPrompt-DE
& 5 & 0 & 35.5 & 0.71 & 36.3 & 0.67 & 35.9$^{\dagger}$ & 0.59 & 42.8 & 1.54 & 20.7 & 1.57 & 27.9$^{\dagger}$ & 1.67 \\
~~~~~+LPO
& 5 & 1 & 35.8 & 0.68 & 40.6 & 0.48 & \textbf{38.0} & 0.56 & 38.8 & 1.37 & 25.6 & 1.89 & 30.9$^{\dagger}$ & 1.79 \\
~~~~~+GreaTer
& 5 & 1 & 35.5 & 0.55 & 37.0 & 0.80 & 36.2$^{\dagger}$ & 0.50 & 38.7 & 1.47 & 28.4 & 2.05 & 32.8$^{\dagger}$ & 1.86 \\
~~~~~+GreaTer-TG
& 5 & 1 & 35.7 & 1.11 & 36.2 & 0.78 & 35.9$^{\dagger}$ & 0.85 & 42.4 & 1.23 & 21.3 & 1.58 & 28.3$^{\dagger}$ & 1.62 \\
~~~~~+GradPO-Gen
& 5 & 1 &  \multicolumn{2}{c}{-} & \multicolumn{2}{c}{-} & \multicolumn{2}{c}{-} & 37.3 & 1.43 & 30.7 & 1.91 & \textbf{33.7} & 1.71 \\
~~~~~+GradPO-Prob
& 5 & 1 & 33.5 & 1.05 & 40.6 & 0.51 & 36.7$^{\dagger}$ & 0.79 & 40.2 & 1.29 & 26.5 & 1.69 & 32.0$^{\dagger}$ & 1.61 \\
\\
EvoPrompt-DE
& 10 & 0 &  \multicolumn{2}{c}{-} & \multicolumn{2}{c}{-} & \multicolumn{2}{c}{-} & 37.2 & 1.33 & 33.3 & 1.89 & 35.2$^{\dagger}$ & 1.64 \\
~~~~~+LPO
& 10 & 1 &  \multicolumn{2}{c}{-} & \multicolumn{2}{c}{-} & \multicolumn{2}{c}{-} & 35.7 & 0.85 & 37.9 & 1.54 & 36.8$^{\dagger}$ & 1.17 \\
~~~~~+GreaTer
& 10 & 1 &  \multicolumn{2}{c}{-} & \multicolumn{2}{c}{-} & \multicolumn{2}{c}{-}& \multicolumn{2}{c}{-} & \multicolumn{2}{c}{-} & \multicolumn{2}{c}{-}\\
~~~~~+GreaTer-TG
& 10 & 1 &  \multicolumn{2}{c}{-} & \multicolumn{2}{c}{-} & \multicolumn{2}{c}{-} & \multicolumn{2}{c}{-} & \multicolumn{2}{c}{-} & \multicolumn{2}{c}{-}\\
~~~~~+GradPO-Gen
& 10 & 1 &  \multicolumn{2}{c}{-} & \multicolumn{2}{c}{-} & \multicolumn{2}{c}{-}& 35.7 & 1.09 & 39.5 & 1.73 & \textbf{37.5} & 1.33 \\
~~~~~+GradPO-Prob
& 10 & 1 &  \multicolumn{2}{c}{-} & \multicolumn{2}{c}{-} & \multicolumn{2}{c}{-} & 37.4 & 1.28 & 33.2 & 1.70 & 35.2$^{\dagger}$ & 1.51 \\




\bottomrule

\end{tabular}
  \caption{Results on \fewrel{} with \rpo{} and \evoprompt{}-DE as first-stage optimizers, and \lpo{}, \greater{}, and \gradpo{} as second-stage refiners.
  R and G denote reasoning-based and local refinement iterations.
  We report P, R, and F1 as mean \(\pm\) std, and bold marks the best F1 within each first-stage block.
  $^{*}$ and $^{\dagger}$ mark statistically significantly better scores within the same block at $p<0.05$ and $p<0.01$.
  Missing values mean the refined prompt did not beat the stage-1 prompt on dev, so test evaluation is omitted.
  For \qwensmall{}, \evoprompt{}-DE at 10 first-stage iterations did not beat the 5-iteration prompt on dev, so we omit it.
  }  
  \label{tab:results_main_fewrel}
\end{table*}

\section{Qualitative Analysis}
\label{sec:qualitative-analysis}

\paragraph{GradPO makes effective local edits with limited impact on prompt readability}
The prompt examples in Appendix~\ref{subsec:optimized-prompts} show that \gradpo{} usually keeps the prompt structure intact and only edits a few local spans.
In the \rpo{} example shown in Figure~\ref{fig:framework}, the \gradpogen{} prompt still reads very similarly to the stage-1 prompt, but the local edits are enough to improve the score.
Overall, the refined prompts remain mostly natural because \gradpo{} only changes a small part of the instruction and aims to keep a low perplexity.
At the same time, second-stage gradient-based local refiners can occasionally introduce minor incoherent surface forms because they operate through small token-level edits.
For example, one edit changes the phrase to ``query inquery sentence,'' which is less natural than the original wording.
Even so, this issue remains local to a small part of the prompt, while the overall prompt structure is preserved.

\paragraph{Most GradPO gains come in the first few iterations.}
\gradpo{} usually improves the prompt in the early iterations, and the first iteration often gives the largest gain.
Later iterations can still help, but the additional improvements are usually small.
This is also consistent with the main results tables, where one \gradpo{} step already improves the stage-1 prompt in most cases.
In Table~\ref{tab:grad_longer_runs}, across the four 10-iteration \gradpo{} runs with \gemmasmall{} and \qwensmall{} on \tacred{} and \fewrel{}, 
two runs reach their best score at the first iteration, one reaches it at the third iteration, and one reaches it at the seventh iteration.
Overall, \gradpo{} works best as a fast local refiner, where most of the benefit comes from the first few updates.


\paragraph{\gradpo{} explores a larger prompt space with limited extra cost compared to baselines.}
The additional computation required by \gradpo{} is small relative to the overall evaluation cost.
Unlike \greater{}, which edits one token at a time, \gradpo{} edits multiple selected spans and therefore searches a richer local prompt space.
In our largest \qwensmall{} setting, we use beam width $B$=$5$, $C$=$5$ candidates per span, and $S$=$5$ spans, producing 105 candidate prompts per iteration.
On the largest training subset of 220 examples, this corresponds to 23,100 training forward calls, 
compared with 5,500 calls for the 25 candidate replacements scored by \greater{}.
This increase is modest because the dominant cost comes from dev-set evaluation: for both \gradpo{} and \greater{}, 
the top 5 prompts are evaluated on 9,000 episodes, each requiring 5 binary LLM calls, for 225,000 forward calls per iteration.
Thus, even in the largest \gradpo{} setting, candidate search accounts for only about 10.3\% of the dev-set evaluation cost, compared with 2.4\% for \greater{}.
\lpo{} also adds optimizer-model calls for error analysis and local rewrites, but these are small compared with repeated train and dev evaluations. 
Overall, \gradpo{} incurs only a modest additional search cost while exploring a substantially richer local prompt space, leading to larger score improvements in our experiments.

\paragraph{GradPO improves prompts without increasing their length}
Refined prompts stay almost the same length because \gradpo{} only edits a few local spans.
We compare 8 matched prompt pairs where \gradpogen{} refines prompts produced by \rpo{}, \evoprompt{}, or \etgpo{}, measuring their lengths with both \qwensmall{} and \gemmasmall{} tokenizers.
Across these pairs, the average length changes from 283.38 to 281.62 tokens for \qwensmall{} and from 290.00 to 288.62 tokens for \gemmasmall{}.
This shows that \gradpo{} does not improve performance by producing longer prompts.
Instead, it preserves the original prompt length and makes only small local edits.

\paragraph{GradPO edits are model-specific and transfer weakly across models}
Table~\ref{tab:cross_model_transfer} shows that cross-model transfer is not very effective for \gradpo{}.
Only 3 out of 8 transferred cases improve when applying the refined prompt to a different target model.
This is expected because \gradpo{} relies on model-specific losses and gradient signals.
As a result, local edits that help one model do not always transfer well to another.

\section{Conclusion}
\label{sec:conclusion}

Few-shot relation extraction with in-context learning still relies heavily on manually written prompts.
This paper studies how to optimize these prompts automatically with a two-stage framework: 
reasoning-based optimization makes broad prompt improvements, while \gradpo{} refines high-impact spans with local gradient-guided edits.
This combination provides consistent second-stage refinement, yields a new state of the art on \tacred{} with \qwensmall, and remains competitive on \fewrel{}.
Overall, our findings show that combining global reasoning-based search with local gradient-based refinement is an 
effective direction for prompt optimization in episodic FSRE.

\pagebreak

\section*{Limitations}

\gradpo{} still depends on several hyperparameters, including the number of editable spans, beam width, and the number of replacement candidates.
Although we tune these values within small ranges, the best settings may vary across models and datasets.
Moreover, the method is not guaranteed to improve every prompt.
In our experiments, most prompts benefit from local refinement, but a small number do not.

Another limitation concerns prompt quality after repeated local edits.
Because \gradpo{} edits only small spans rather than rewriting the full prompt, it can sometimes introduce minor local inconsistencies.
We also evaluate only smaller open models, namely \qwensmall{} and \gemmasmall{}, in the main experiments.
As a result, it remains unclear how the same two-stage setup would behave for larger LLMs.

Finally, our study focuses on a non-reasoning task and is limited to FSRE.
Future work could extend this framework beyond FSRE and explore reasoning-based tasks.

\bibliography{custom}

\appendix

\section{Additional details and hyperparameters of the optimization methods}
\label{sec:additional_details}

\begin{table*}
  \centering
  \small
  \setlength{\tabcolsep}{.061in}
\begin{tabular}{l r@{+}l
                r@{ \scriptsize$\pm$ }l r@{ \scriptsize$\pm$ }l l@{ \scriptsize$\pm$ }l
                r@{ \scriptsize$\pm$ }l r@{ \scriptsize$\pm$ }l l@{ \scriptsize$\pm$ }l}
\toprule
  &
  \multicolumn{2}{c}{\#itr.} &
  \multicolumn{6}{c}{Qwen3-4B} &
  \multicolumn{6}{c}{Gemma3-4B} \\ \cmidrule(lr){4-9} \cmidrule(lr){10-15}
& R & G &
  \multicolumn{2}{c}{P} & \multicolumn{2}{c}{R} & \multicolumn{2}{c}{F1} &
  \multicolumn{2}{c}{P} & \multicolumn{2}{c}{R} & \multicolumn{2}{c}{F1} \\ \midrule

\multicolumn{8}{l}{Performance on \tacred} \\
\midrule
Initial Prompt             & \multicolumn{2}{c}{} & 17.6 & 0.44 & 38.2 & 1.38 & 24.0 & 0.59 & 10.4 & 0.56 & 12.5 & 0.65 & 11.3 & 0.53 \\ 
\\
ETGPO
& 1 & 0 & 21.5 & 0.54 & 41.4 & 1.80 & 28.3 & 0.75 & 12.1 & 0.43 & 14.4 & 0.36 & 13.1$^{\dagger}$ & 0.32 \\
~~~~~+LPO
& 1 & 1 & 21.3 & 0.73 & 42.3 & 1.56 & 28.3 & 0.89 & 11.0 & 0.40 & 14.4 & 0.62 & 12.5$^{\dagger}$ & 0.41 \\
~~~~~+GreaTer
& 1 & 1 & 21.2 & 0.56 & 43.2 & 1.32 & \textbf{28.4} & 0.67 &  \multicolumn{2}{c}{-} & \multicolumn{2}{c}{-} & \multicolumn{2}{c}{-}\\
~~~~~+GreaTer-TG
& 1 & 1 & 27.5 & 0.87 & 28.1 & 1.27 & 27.8 & 0.92 &  \multicolumn{2}{c}{-} & \multicolumn{2}{c}{-} & \multicolumn{2}{c}{-}\\
~~~~~+GradPO-Gen
& 1 & 1 & 22.0 & 0.49 & 37.3 & 1.55 & 27.7$^{\dagger}$ & 0.66 & 11.5 & 0.51 & 18.2 & 0.77 & \textbf{14.1} & 0.58 \\
~~~~~+GradPO-Prob
& 1 & 1 & 24.1 & 1.17 & 29.0 & 0.91 & 26.3$^{\dagger}$ & 1.03 &  \multicolumn{2}{c}{-} & \multicolumn{2}{c}{-} & \multicolumn{2}{c}{-}\\
\midrule
\multicolumn{8}{l}{Performance on \fewrel} \\ 
\midrule
Initial Prompt             & \multicolumn{2}{c}{} & 29.9 & 0.72 & 44.1 & 0.76 & 35.6 & 0.66 & 39.9 & 2.13 & 18.6 & 1.12 & 25.3 & 1.27 \\
\\
ETGPO
& 1 & 0 & 35.6 & 1.35 & 34.5 & 0.93 & 35.0$^{\dagger}$ & 1.06 & 33.8 & 1.50 & 36.6 & 2.14 & \textbf{35.2} & 1.73 \\
~~~~~+LPO
& 1 & 1 & \multicolumn{2}{c}{-} & \multicolumn{2}{c}{-} & \multicolumn{2}{c}{-} & 35.9 & 1.96 & 22.7 & 1.30 & 27.8$^{\dagger}$ & 1.52 \\
~~~~~+GreaTer
& 1 & 1 & 36.2 & 1.44 & 35.1 & 1.21 & \textbf{35.7} & 1.24 &  \multicolumn{2}{c}{-} & \multicolumn{2}{c}{-} & \multicolumn{2}{c}{-}\\
~~~~~+GreaTer-TG
& 1 & 1 & \multicolumn{2}{c}{-} & \multicolumn{2}{c}{-} & \multicolumn{2}{c}{-} & 35.8 & 1.51 & 34.2 & 2.08 & 35.0 & 1.76\\
~~~~~+GradPO-Gen
& 1 & 1 & 34.3 & 1.22 & 36.0 & 0.95 & 35.1$^{\dagger}$ & 1.04 &  \multicolumn{2}{c}{-} & \multicolumn{2}{c}{-} & \multicolumn{2}{c}{-}\\
~~~~~+GradPO-Prob
& 1 & 1 & 33.8 & 1.14 & 36.8 & 0.81 & 35.2$^{*}$ & 0.93 &  \multicolumn{2}{c}{-} & \multicolumn{2}{c}{-} & \multicolumn{2}{c}{-}\\
\bottomrule

\end{tabular}
  \caption{Results with \etgpo{} as the first-stage optimizer on \tacred{} and \fewrel{}, followed by \lpo{}, \greater{}, and \gradpo{} as second-stage refiners.
  R and G denote reasoning-based and local refinement iterations, respectively.
  We report precision, recall, and F1 as mean \(\pm\) standard deviation, and bold marks the best F1 within each dataset/model block.
  $^{*}$ and $^{\dagger}$ mark scores that are statistically significantly better within the same stage-1 block at $p<0.05$ and $p<0.01$, respectively.
  Missing values indicate that the refined prompt did not improve over the \etgpo{} prompt on the dev set, so we omit test evaluation.
  }  
  \label{tab:results_etgpo}
\end{table*}

\begin{table*}[t]
  \centering
  \small
\begin{tabular}{l 
                r@{ \scriptsize $\pm$ }l r@{ \scriptsize $\pm$ }l r@{ \scriptsize $\pm$ }l
                r@{ \scriptsize $\pm$ }l r@{ \scriptsize $\pm$ }l r@{ \scriptsize $\pm$ }l}
\toprule
  &
  \multicolumn{6}{c}{\tacred} \\ \cmidrule(lr){2-7}
& 
  \multicolumn{2}{c}{P} & \multicolumn{2}{c}{R} & \multicolumn{2}{c}{F1} \\ \midrule

MNAV~\cite{fs-tacred}        & \multicolumn{2}{c}{-} & \multicolumn{2}{c}{-} & 12.4 & {\scriptsize 1.01}   \\

OdinSynth~\cite{odinsynth}   & 23.5 & {\scriptsize 1.46} & 11.5 & {\scriptsize 1.02} & 15.4 & {\scriptsize 1.21} \\
                     
CKPT~\cite{CKPT}             & \multicolumn{2}{c}{-} & \multicolumn{2}{c}{-} & 15.1 & {\scriptsize 1.12}  \\

Anchor+gen. rules~\cite{anchorword} & 19.6 & {\scriptsize 0.63} & 31.9 & {\scriptsize 1.04} & 24.2 & {\scriptsize 0.72} \\

SoftRules~\cite{softmatcher}  & 33.5 & {\scriptsize 1.47} & 19.7 & {\scriptsize 1.14} & 24.8 & {\scriptsize 1.22} \\                            
ICL with \qwensmall\ \cite{struct-icl} & 26.4 & {\scriptsize 0.69} & 28.9 & {\scriptsize 1.88} & {27.6} & {\scriptsize {1.20}} \\ 
RPO+GradPO with \qwensmall\  (Ours) & 29.5 & {\scriptsize 1.25} & 28.5 & {\scriptsize 1.66} & \textbf{29.0} & {\scriptsize 1.37} \\
\bottomrule

\end{tabular}

  \caption{Comparison of previous work on \tacred{} with our best prompt optimization setting using \qwensmall{}.
  All methods use the same episodic \tacred{} benchmark in the $1$-shot setting.
  With \qwensmall{}, our two-stage prompt optimization approach improves over the previous best result reported with the same model size.
  }  
  \label{tab:results_tacred_revious_works}
\end{table*}

\begin{table*}[t]
\centering
\small
\setlength{\tabcolsep}{4pt}
\begin{tabular}{l r@{--}l r@{ \scriptsize$\pm$ }l c r@{ \scriptsize$\pm$ }l c}
\toprule
& \multicolumn{2}{c}{itr.} & \multicolumn{3}{c}{\qwensmall} & \multicolumn{3}{c}{\gemmasmall} \\
\cmidrule(lr){2-3} \cmidrule(lr){4-6} \cmidrule(lr){7-9}
Method & \multicolumn{2}{c}{range} & \multicolumn{2}{c}{Avg.\ F1} & {Best F1} & \multicolumn{2}{c}{Avg.\ F1} & {Best F1} \\
\midrule
\multicolumn{8}{l}{Performance on \tacred} \\
\midrule
\rpo & 1 & 10 & 26.0 & 3.39 & 30.3 & 9.0 & 3.79 & 13.8  \\
\rpo & 11 & 20 & 26.1 & 4.46 & 33.6 & 9.4 & 3.79 & 16.5 \\
\rpo + 1-step \gradpo{-Gen} & \multicolumn{2}{c}{after 10} & \multicolumn{2}{c}{-} & 31.3 & \multicolumn{2}{c}{-} & 17.1  \\
\\
\evoprompt & 1 & 10 & 21.0 & 2.56 & 23.1  & 13.6 & 3.07 & 19.0 \\
\evoprompt & 11 & 20 & 22.0 & 0.68 & 24.6  & 10.4 & 0.76 & 18.0  \\
\evoprompt + 1-step \gradpo{-Gen} & \multicolumn{2}{c}{after 10} & \multicolumn{2}{c}{-} & 23.5  & \multicolumn{2}{c}{-} & 20.3 \\

\midrule
\multicolumn{8}{l}{Performance on \fewrel} \\
\midrule
\rpo & 1 & 10 & 22.3 & 3.02 & 30.8  & 14.9 & 4.62 &  15.8  \\
\rpo & 11 & 20 & 23.9 & 3.89 & 31.1 & 15.0 & 3.01 & 17.7  \\
\rpo + 1-step \gradpo{-Gen} & \multicolumn{2}{c}{after 10} & \multicolumn{2}{c}{-} & 31.5 & \multicolumn{2}{c}{-} & 19.7  \\
\\
\evoprompt & 1 & 10 & 20.8 & 3.09 & 24.7 & 12.8 & 5.92 & 18.5  \\
\evoprompt & 11 & 20 & 22.3 & 1.72 & 25.1 & 15.0 & 4.75 & 19.7  \\
\evoprompt + 1-step \gradpo{-Gen} & \multicolumn{2}{c}{after 10} & \multicolumn{2}{c}{-} & 23.3 & \multicolumn{2}{c}{-} & 21.0  \\

\bottomrule
\end{tabular}

\caption{Dev-set analysis of later iterations of reasoning-based prompt optimization and a one-step \gradpogen{} refinement after iteration 10.
Later reasoning-based iterations still show slight average improvements, but their prompt quality remains unstable across iterations.
A single \gradpogen{} step, which searches a larger local edit space, usually provides an additional stable improvement over an already strong stage-1 prompt.}
\label{tab:gradpo_stability}
\end{table*}

\begin{table*}
\centering
\small
\setlength{\tabcolsep}{.045in}
\begin{tabular}{l r@{+}l r@{ \scriptsize$\pm$ }l r@{ \scriptsize$\pm$ }l r@{ \scriptsize$\pm$ }l r@{ \scriptsize$\pm$ }l r@{ \scriptsize$\pm$ }l r@{ \scriptsize$\pm$ }l}
\toprule
& \multicolumn{2}{c}{\#itr.} & \multicolumn{6}{c}{\qwensmall\ (Optimized on \gemmasmall)} & \multicolumn{6}{c}{\gemmasmall\ (Optimized on \qwensmall)} \\
\cmidrule(lr){4-9} \cmidrule(lr){10-15}
Method & R & G & \multicolumn{2}{c}{P} & \multicolumn{2}{c}{R} & \multicolumn{2}{c}{F1} & \multicolumn{2}{c}{P} & \multicolumn{2}{c}{R} & \multicolumn{2}{c}{F1} \\
\midrule
RPO & 5 & 0 & 24.4 & 0.66 & 30.9 & 1.05 & 27.2 & 0.68 & 14.7 & 1.31 & 04.6 & 0.51 & 07.0 & 0.70 \\
~~~+GradPO-Gen & 5 & 1 & 24.0 & 0.62 & 30.4 & 1.69 & 26.8 & 0.97 & 12.9 & 0.41 & 06.2 & 0.52 & 08.4 & 0.55 \\
RPO & 10 & 0 & 17.5 & 0.68 & 35.2 & 2.02 & 23.3 & 1.03 & 19.5 & 1.65 & 02.9 & 0.36 & 05.0 & 0.59 \\
~~~+GradPO-Gen & 10 & 1 & 15.3 & 0.39 & 41.1 & 1.62 & 22.3 & 0.62 & 17.1 & 1.25 & 03.4 & 0.26 & 05.7 & 0.42 \\

\midrule
RPO & 5 & 0 & 25.6 & 0.66 & 51.1 & 0.93 & 34.1 & 0.74 & 43.9 & 3.18 & 12.1 & 1.15 & 19.0 & 1.60 \\
~~~+GradPO-Gen & 5 & 1 & 17.6 & 0.67 & 56.5 & 1.00 & 26.9 & 0.87 & 44.5 & 2.53 & 13.6 & 0.94 & 20.8 & 1.27 \\
RPO & 10 & 0 & 27.3 & 0.86 & 52.6 & 1.50 & 36.0 & 1.00 & 46.2 & 3.43 & 07.1 & 0.96 & 12.3 & 1.56 \\
~~~+GradPO-Gen & 10 & 1 & 23.6 & 1.00 & 54.9 & 2.08 & 33.0 & 1.30 & 46.6 & 4.27 & 05.8 & 0.91 & 10.3 & 1.53\\
\bottomrule
\end{tabular}

\caption{Cross-model transfer evaluation on \tacred{}, where prompts are optimized on one model and evaluated on the other model.
R and G denote the numbers of reasoning-based and gradient-based iterations, respectively.
The transfer gains are limited, which suggests that \gradpo{} edits are largely model-specific.
This is expected because the refinements depend on the target model's loss and gradient signals.}
\label{tab:cross_model_transfer}
\end{table*}

\begin{table*}
\centering
\small
\setlength{\tabcolsep}{.045in}
\setlength{\tabcolsep}{.061in}
\begin{tabular}{l r@{+}l
                r@{ \scriptsize$\pm$ }l r@{ \scriptsize$\pm$ }l l@{ \scriptsize$\pm$ }l
                r@{ \scriptsize$\pm$ }l r@{ \scriptsize$\pm$ }l l@{ \scriptsize$\pm$ }l}
\toprule
  &
  \multicolumn{2}{c}{\#itr.} &
  \multicolumn{6}{c}{Qwen3-4B} &
  \multicolumn{6}{c}{Gemma3-4B} \\ \cmidrule(lr){4-9} \cmidrule(lr){10-15}
& R & G &
  \multicolumn{2}{c}{P} & \multicolumn{2}{c}{R} & \multicolumn{2}{c}{F1} &
  \multicolumn{2}{c}{P} & \multicolumn{2}{c}{R} & \multicolumn{2}{c}{F1} \\ \midrule
\multicolumn{8}{l}{Performance on \tacred} \\
\midrule
Initial Prompt             & \multicolumn{2}{c}{} & 17.6 & 0.44 & 38.2 & 1.38 & 24.0 & 0.59 & 10.4 & 0.56 & 12.5 & 0.65 & 11.3 & 0.53 \\ 
\\
\rpo
& 10 & 0 & 35.8 & 1.19 & 23.4 & 1.26 & 28.3 & 1.13 & 09.8 & 0.33 & 14.4 & 0.72 & 11.7 & 0.45 \\
\evoprompt
& 10 & 0 & 21.6 & 0.53 & 28.4 & 1.77 & 24.5 & 0.98 & 12.5 & 0.54 & 08.5 & 0.48 & 10.1 & 0.42 \\
\etgpo
& 10 & 0 & 21.5 & 0.54 & 41.4 & 1.80 & 28.3 & 0.75 & 12.1 & 0.43 & 14.4 & 0.36 & 13.1 & 0.32 \\
\greater
& 0 & 10 & 20.5 & 0.81 & 28.9 & 2.07 & 24.0 & 1.21 & 09.0 & 0.42 & 11.1 & 0.65 & 09.9 & 0.43 \\
\gradpogen
& 0 & 10 & 17.5 & 0.46 & 40.3 & 1.17 & 24.4 & 0.52 & 09.8 & 0.53 & 10.2 & 0.66 & 10.0 & 0.57 \\
\midrule
\multicolumn{8}{l}{Performance on \fewrel} \\
\midrule
Initial Prompt             & \multicolumn{2}{c}{} & 29.9 & 0.72 & 44.1 & 0.76 & 35.6 & 0.66 & 39.9 & 2.13 & 18.6 & 1.12 & 25.3 & 1.27 \\
\\
\rpo
& 10 & 0 & 33.5 & 0.98 & 39.6 & 0.66 & 36.3 & 0.74 & 34.8 & 1.39 & 29.9 & 1.42 & 32.2 & 1.30 \\
\evoprompt
& 10 & 0 & 35.5 & 0.71 & 36.3 & 0.67 & 35.9 & 0.59  & 37.2 & 1.33 & 33.3 & 1.89 & 35.2 & 1.64 \\
\etgpo
& 10 & 0 & 35.6 & 1.35 & 34.5 & 0.93 & 35.0 & 1.06 & 33.8 & 1.50 & 36.6 & 2.14 & 35.2 & 1.73 \\
\greater
& 0 & 10 & 33.1 & 0.85 & 39.8 & 1.04 & 36.2 & 0.85 & 34.7 & 1.10 & 29.7 & 1.63 & 32.0 & 1.39 \\
\gradpogen
& 0 & 10 & 32.1 & 0.78 & 43.7 & 0.38 & 37.0 & 0.63 & 37.3 & 0.99 & 29.2 & 1.39 & 32.7 & 1.17 \\
\bottomrule

\end{tabular}
\caption{Results with longer runs of gradient-based prompt optimization on \tacred{} and \fewrel{}.
R and G denote the numbers of reasoning-based and gradient-based iterations, respectively.
This table shows how much the gradient-based methods improve when they are allowed to run for more iterations after starting from the same initial prompt.}
\label{tab:grad_longer_runs}
\end{table*}

\subsection{Additional details on optimizer methods}
\label{sec:additional_details_opt}

\paragraph{\rpo}
At each iteration, we sample $100$ training pairs, each consisting of a relation with one support sentence and a query, 
and retain three feedback examples using mixed selection, which includes at least one correct and one incorrect prediction when available.
Since the validation set is relatively small, prompt scores can fluctuate noticeably across runs. To favor more stable prompts during selection, we penalize the standard deviation and use the score
\begin{equation}
\mathrm{F1}_{\mathrm{mean}} - 2 \times \mathrm{F1}_{\mathrm{std}}.
\label{eq:stable_score}
\end{equation}

Parent prompts are sampled with a softmax distribution over this score using temperature $1.0$, and limit the population size at $10$.

\paragraph{\evoprompt}
For \evoprompt{}, we use the DE variant with a population size of five.
The initial population contains the two human written prompt and three AI written prompt variants.
At each iteration, we sample $1,000$ training episodes, and score the prompts on it using Equation~\ref{eq:stable_score}.
Only the best prompt from the population is evaluated on the validation set at the end of each iteration.

\paragraph{\etgpo}
The original method generates a single prompt from the error taxonomy.
Instead, we generate five prompts from the same error taxonomy and evaluate them on the validation set to select the best one.
The error taxonomy is constructed from errors observed on $1,000$ sampled training pairs.
We use the default hyperparameters: taxonomy batch size $6$, coverage threshold $0.7$, and minimum category size $2$.
We tune the maximum number of error categories between $5$ and $7$, and select $5$ based on validation performance.

\paragraph{\gradpo}
For \gradpo{}, we tune the training gradient sample size between 1,000 and 50,000 pairs, and use 20,000 pairs for \qwensmall{} and 30,000 pairs for \gemmasmall{}.
Since more than 90\% of the training labels are \texttt{no\_relation}, we construct a balanced gradient set from the sampled pairs by taking equal numbers of TP, TN, FP, and FN instances, limited by the smallest group.
This reduces the bias toward \texttt{no\_relation} and yields more balanced gradient signals.

We tune the number of candidate replacements per span between 3 and 7, selecting 5 for both \qwensmall{} and \gemmasmall{}.
We also tune the number of edited spans between 3 and 7, selecting 5 for \qwensmall{} and 3 for \gemmasmall{}.
The span expansion ratio is tuned between 0.3 and 0.7, with 0.6 performing best for both models.
We keep the beam width fixed at 5.
We tune the perplexity penalty between 0.2 and 1.0, and use 0.5 for both models. 
Following \greater{}, we use the log-perplexity form rather than the exponential perplexity value.
The maximum span length is tuned between 1 and 5 tokens.
After evaluating the beam candidates on the validation set, we select the best prompt using Equation~\ref{eq:stable_score}, 
which is then used as the starting prompt for the next iteration.

For \gradpoprob{}, we provide context about the full prompt; otherwise, when suggesting tokens for the spans near the beginning of the prompt,
the model often generates irrelevant replacement tokens due to lack of context. 
Specifically, we prepend the original prompt with an instruction to paraphrase it as additional context.

Moreover, we restrict span selection to preserve task-specific terminology used consistently across the static instruction and episode-specific input fields. 
In particular, spans containing \texttt{relation}, \texttt{query}, \texttt{support}, \texttt{yes}, or \texttt{no} are split so that these tokens are not edited. 
Although allowing edits to these terms improves development scores even more, 
it may introduce incoherent prompts or inconsistencies with the fixed input template; 
therefore, we keep these terms unchanged to preserve prompt fluency and consistency. 
For each selected span, we also include the original span as a candidate, so beam search can leave it unchanged when needed.

\paragraph{\greater}
For consistency with \gradpo{}, we use the same training sample sizes: 20,000 pairs for \qwensmall{} and 30,000 pairs for \gemmasmall{}.
We also use the same balanced subset construction to reduce the bias toward \texttt{no\_relation}, taking equal numbers of TP, TN, FP, and FN instances from the sampled pairs, limited by the smallest group size.
Gradients and training losses are computed on this balanced subset.

We set the proposal top-\(k\) to 25 and top-\(u\) to 5.
At the end of each iteration, we evaluate the top-\(u\) prompts on the validation set and keep the resulting snapshots, while the training loop retains the best prompt according to Equation~\ref{eq:local_refinement_loss}.
We tune the fluency weight between 0.0 and 1.0, and find that 0.2 performs best on the validation set.

Moreover, in our relation extraction inference templates, episode-specific input details are appended after the instruction prompt. 
In contrast, \greater{} prepends the input details in its original prompting setup before generating reasoning and collecting token probabilities. 
To adapt \greater{} to our task, we provide input context for collecting token suggestions by 
prepending a simple scenario description with a sample input, similar to \gradpoprob{}.

\paragraph{\lpo}
For \lpo{}, we use the same balanced subset construction as \gradpo{} and \greater{}.
At each step, we provide the optimizer with 3 examples ofincorrect predictions.
We tune the number of editable spans between 3 and 5, with each span limited to at most 3 words in the meta-prompt.
We generate 5 local rewrites and evaluate all of them directly on the validation set using Equation~\ref{eq:stable_score}.

\subsection{Dataset split details}
\label{sec:dataset_split_details}

For \tacred{}, the train sets and validation sets are created from the corresponding original train and validation splits.
For \fewrel{}, the train sets and validation sets are created from the original FewRel train split, because the test episodes are created from the original validation split.
The original FewRel train split contains $64$ unique relations, which we divide into $54$ train relations and $10$ validation relations.
For both datasets, we create the validation set as three sets of $3{,}000$ episodes.

\subsection{Entity-match Filtering}
Following the work by~\citet{struct-icl}, we employ entity-match filtering during test inference.
Before relation matching, we prompt the model to check whether the subject and object entities match the entity types required by the candidate relation.
If this check fails, we skip relation matching for that candidate and directly assign a negative prediction.
As shown in prior work, this filtering improves F1 by reducing invalid positive predictions.

\subsection{Decoding and inference Details}
For relation inference and entity-type checking, we formulate prediction as binary classification using the logits of the first generated token, restricted to ``yes'' and ``no''.
We predict ``yes'' when \( P(\texttt{yes}) \geq P(\texttt{no}) \), and ``no'' otherwise.

For generation-based components, we use the default decoding configuration of each model.
For Qwen models, we use sampling with temperature $0.6$, top-\(p=0.95\), and top-\(k=20\).
For Gemma models, we use sampling with temperature $1.0$, top-\(p=0.95\), and top-\(k=64\).
We do not truncate input prompts, and set the max tokens for generation to $10{,}000$ for any prompting calls.

\section{Input Prompt Format}
\label{sec:prompt_template}

We construct the full input prompt as three components: the \texttt{Instruction Prompt}, 
the \texttt{Answer Instruction Prompt}, and the \texttt{Input Details Prompt}.
The \texttt{Instruction Prompt} defines the task and decision guidelines,
the \texttt{Answer Instruction Prompt} specifies the required output format in one simple sentence,
and the \texttt{Input Details Prompt} contains the episode-specific inputs such as the relation name,
relation description, support sentence, and query sentence.

During optimization, we modify only the \texttt{Instruction Prompt}.
At inference time, the final prompt is formed by concatenating the three components with newline separators:

\begin{flushleft}
\verb|```|\\
\texttt{\{Instruction Prompt\}} \\[0.2em]
\texttt{\{Answer Instruction Prompt\}} \\[0.2em]
\texttt{\{Input Details Prompt\}}\\
\verb|```|
\end{flushleft}

The complete set of prompts used in our framework, including input prompts, meta-prompts, 
and optimized prompts, is provided in Appendix~\ref{sec:prompts}.

\newpage

{\onecolumn

\section{Prompts}
\label{sec:prompts}

\subsection{Input Prompt Templates}
\label{subsec:input_prompt_template_details}

\begin{tcolorbox}[promptbox={Instruction prompt (the initial prompt which will be optimized)}]
\small
\ttfamily

You are given a relation name, a description of the relation in brackets, a support sentence exemplifying the relation, and a query sentence.\\

A relation connects the Subject and the Object entities. The Subject and the Object entities are indicated with the subject and object tags, respectively.

You need to decide whether the relation holds between the Subject and the Object entities in the query sentence.\\

If the relation holds between the Subject and the Object entities in the query sentence, answer "yes"; otherwise, answer "no".

\end{tcolorbox}

\begin{tcolorbox}[promptbox={Answer instruction prompt}]
\small
\ttfamily

Output only "yes" or "no" as answer, with no explanation or additional text.

\end{tcolorbox}

\begin{tcolorbox}[promptbox={Input prompt}]
\small
\ttfamily

==\\

Relation name: "\#RELATION\#" (\#RELATION\_DESCRIPTION\#)\\

Support sentence: \#SUPPORT\_SENTENCE\#\\

Query Sentence: \#QUERY\_SENTENCE\#\\

Answer: 
\end{tcolorbox}

\subsection{Meta-prompts}
\label{subsec:meta-prompts}

\subsubsection{\gradpo{} meta-prompts}
\label{subsec:gradpo-meta-prompts}

\begin{tcolorbox}[promptbox={Replacement generation prompt}]
\small
\ttfamily
You are an expert on suggesting replacements for targeted spans in a prompt. \\
\\
The editable spans are marked using tags such as <span\_1>...</span\_1>, <span\_2>...</span\_2>, and so on. \\
\\
Input Prompt with editable spans: \\
\verb|```| \\
\#MARKED\_PROMPT\# \\
\verb|```| \\
\\
Editable spans: \\
\#REGION\_CANDIDATE\_REQUEST\_BLOCKS\# \\
\\
Task: \\
Suggest \#NUM\_CANDIDATES\# replacements for each editable span so that every replacement fits naturally in context. \\
\\
Rules: \\
- Preserve the meaning and role of each editable span \\
- Treat each tagged span as a single unit \\
- Return replacements only for the editable spans \\
- A candidate may be identical to the original span text if keeping it unchanged is the best option \\
\\
Please reason through the problem, but output the replacements in JSON: \\
\verb|```|json \\
\{ \\
  ``span\_1'': \{ \\
    ``candidates'': [ \\
      ``replacement 1'', \\
      ``...'', \\
      ``replacement \#NUM\_CANDIDATES\#'' \\
    ] \\
  \}, \\
  ... \\
  ``span\_\#NUM\_REGIONS\#'': \{ \\
    ``candidates'': [ \\
      ``replacement 1'', \\
      ``...'', \\
      ``replacement \#NUM\_CANDIDATES\#'' \\
    ] \\
  \} \\
\} \\
\verb|```| \\
\end{tcolorbox}

\begin{tcolorbox}[promptbox={Span replacement prompt}]
\small
\ttfamily
You are an expert prompt generator for a relation extraction inference task. \\
\\
A relation captures the connection between two entities in a sentence by describing their relationship. We will refer to these entities as the subject and object entities. \\
The task requires inferring a binary (yes/no) answer based on whether the query sentence expresses this relation between the subject and the object entities. \\
\\
You are given the current instruction prompt with targeted spans below: \\
\verb|```| \\
\#ALL\_MARKED\_PROMPT\# \\
\verb|```| \\
Each editable span is marked with tags such as <span\_1>...</span\_1>, <span\_2>...</span\_2>, and so on. \\
\\
Replacements for each spans are given below: \\
\#SELECTED\_REPLACEMENTS\# \\
\\
Your task is to generate a revised instruction prompt by applying the given replacements to the corresponding spans. \\
Use the replacements exactly as provided, except for minimal local adjustments if necessary for spelling and grammatical correctness or coherence. \\
Do not modify any other parts of the prompt (but remove the span tags). \\
\\
Output only the revised prompt. \\
\end{tcolorbox}

\begin{tcolorbox}[promptbox={Additional context prepended for \gradpoprob{} token generation}]
\small
\ttfamily
Original Prompt: \\
\verb|```|\#FULL\_INSTRUCTION\_PROMPT\#\verb|```| \\
\\
Task: \\
Write a natural revised version of the original prompt while preserving meaning, structure, and tone. Prefer paraphrase or other clear, robust, and effective wording. \\
\\
Revised prompt: \\
\verb|```|\#PROMPT\_PREFIX\_TEXT\# \\
\end{tcolorbox}

\subsubsection{\lpo{} meta-prompts}
\label{subsec:lpo-meta-prompts}

\begin{tcolorbox}[promptbox={Span identification prompt}]
\small
\ttfamily
A relation extraction prompt helps an LLM decide whether a query sentence expresses a target relation between the subject and object entities, using a support sentence for that relation. The classifier must answer with exactly one token: ``yes'' or ``no''. \\
\\
Current prompt: \\
\verb|```| \\
\#INFERENCE\_PROMPT\# \\
\verb|```| \\
\\
Feedback examples from the current prompt: \\
\#FEEDBACK\_EXAMPLES\# \\
\\
First, think about the issues with the prompt. \\
Identify the scope of tokens within the prompt where edits should take place. \\
Prompt edits include adding, deleting or modifying tokens. \\
Mark the scope of the prompt that needs editing by putting <edit>, </edit> tags. \\
You can have multiple <edit> tags and each <edit> tag should not entail more than \#MAX\_WORDS\_PER\_EDIT\_TAG\# words. \\
Use at most \#MAX\_EDIT\_TAGS\# <edit> tags. \\
Do not cover the whole sentence with multiple <edit> tags. \\
\\
Reply with the full prompt enclosed within <p> and </p> tags, with the edit locations marked using <edit> and </edit> tags. \\
Do not include any other text. \\
\end{tcolorbox}

\begin{tcolorbox}[promptbox={Span replacement prompt}]
\small
\ttfamily
A relation extraction prompt helps an LLM decide whether a query sentence expresses a target relation between the subject and object entities, using a support sentence for that relation. The classifier must answer with exactly one token: ``yes'' or ``no''. \\
\\
Current prompt with local edit scopes: \\
\verb|```| \\
\#TAGGED\_PROMPT\# \\
\verb|```| \\
\\
Feedback examples from the current prompt: \\
\#FEEDBACK\_EXAMPLES\# \\
\\
Generate one revised full prompt by editing only the text inside the <edit>, </edit> tags. \\
You may add, delete, or modify tokens inside those local scopes. \\
Keep the rest of the prompt unchanged except for minimal grammar cleanup at the edit boundaries. \\
Keep the <edit> and </edit> tags around the edited local scopes in the revised prompt. \\
\\
Please reason through the problem, but output only the revised prompt enclosed within the <p> and </p> tags. \\
\end{tcolorbox}

\subsubsection{\greater{} meta-prompts}
\label{subsec:greater-meta-prompts}

\begin{tcolorbox}[promptbox={Additional prepended context for token generation}]
\small
\ttfamily
You are optimizing an instruction prompt for a binary relation extraction  classifier. \\
\\
The instruction prompt will appear before the classifier input. \\
Later, the classifier will receive an input similar to the following example and must decide whether the target relation is expressed between the Subject and Object entities in the query sentence. \\
\\
The classifier must answer with exactly one token: ``yes'' or ``no''. \\
\\
Example classifier input: \\
\\
Relation name: \#RELATION\# (\#RELATION\_DESCRIPTION\#) \\
Support Sentence: \#SUPPORT\_SENTENCE\# \\
Query Sentence: \#QUERY\_SENTENCE\# \\
\\
Write a instruction prompt that should appear before this type of input and help the classifier solve the relation extraction problem. \\
\\
Instruction: \\
\end{tcolorbox}

\subsubsection{\etgpo{} meta-prompts}
\label{subsec:etgpo-meta-prompts}

\begin{tcolorbox}[promptbox={First batch taxonomy creation prompt}]
\small
\ttfamily
You are an expert at analyzing why language models fail on relation extraction tasks with binary yes/no inference. \\
\\
\#FAILURES\_TEXT\# \\
\\
\#\# Your Task \\
\\
Analyze each failure and identify the root cause of each error. Focus on the underlying failure mechanism. \\
Important: the reasoning field is post-hoc feedback describing the most likely cause of the incorrect binary prediction, not a verbatim chain-of-thought from the original model. Use it as probabilistic evidence together with the input, gold label, and wrong answer. \\
\\
For each failure, find: \\
1. The EARLIEST point in the reasoning where something went wrong \\
2. What specifically went wrong (calculation error, wrong approach, misunderstanding, etc.) \\
3. Why this error led to the wrong final answer \\
\\
Create issue categories that capture each type of error. Categories should be general enough to potentially apply to other traces, but specific enough to be meaningful, without becoming tied to relation- or example-specific details. \\
\\
IMPORTANT: Each category must be SELF-CONTAINED and understandable by someone who has NOT seen the original problems. \\
\\
\#\# Output Format \\
\\
Return a JSON object with: \\
\\
\verb|```|json \\
\{ \\
    ``categories'': [ \\
        \{ \\
            ``category\_name'': ``Short descriptive name for this type of error'', \\
            ``summary'': ``One sentence describing the core error pattern.'', \\
            ``description'': ``Describe the general failure mechanism in this category.'', \\
            ``example'': ``A concrete, self-contained example. Format: 'Problem: [simple problem]. Error: [what the model does wrong]. Correct: [what should happen].''', \\
            ``error\_type'': ``Type of error (e.g., Calculation Error, Wrong Approach, Conceptual Misunderstanding, Missing Step, Logical Fallacy, Factual Error, Incomplete Reasoning, Misreading the Problem)'', \\
            ``why\_leads\_to\_wrong\_answer'': ``Explanation of how this error causes wrong answers'' \\
        \} \\
    ], \\
    ``failure\_assignments'': [ \\
        \{ \\
            ``failure\_id'': 1, \\
            ``problem\_idx'': <problem\_idx>, \\
            ``run\_id'': <run\_id>, \\
            ``category\_name'': ``Name of the category this failure belongs to'', \\
            ``trace\_details'': \{ \\
                ``trace\_specific\_location'': ``Where in the reasoning the error occurred'', \\
                ``trace\_specific\_details'': ``Specific details about what went wrong'' \\
            \} \\
        \} \\
    ] \\
\} \\
\verb|```| \\
\end{tcolorbox}

\begin{tcolorbox}[promptbox={Subsequent batch taxonomy creation prompt}]
\small
\ttfamily
You are an expert at analyzing why language models fail on relation extraction tasks with binary yes/no inference. \\
\\
\#\# Existing Issue Categories \\
\\
\#EXISTING\_CATEGORIES\_TEXT\# \\
\\
\#FAILURES\_TEXT\# \\
\\
\#\# Your Task \\
\\
For each failure: \\
1. Determine if the error fits one of the EXISTING categories \\
2. OR create a NEW category if the error is fundamentally different \\
Important: the reasoning field is post-hoc feedback describing the most likely cause of the incorrect binary prediction, not a verbatim chain-of-thought from the original model. Use it as probabilistic evidence together with the input, gold label, and wrong answer. \\
\\
New categories should stay general and reusable, rather than becoming tied to relation- or example-specific details. \\
\\
\#\# Output Format \\
\\
Return a JSON object with: \\
\\
\verb|```|json \\
\{ \\
    ``new\_categories'': [ \\
        \{ \\
            ``category\_name'': ``Short descriptive name for NEW error type'', \\
            ``summary'': ``One sentence describing the core error pattern.'', \\
            ``description'': ``Describe the general failure mechanism in this category.'', \\
            ``example'': ``A concrete example.'', \\
            ``error\_type'': ``Type of error (e.g., Calculation Error, Wrong Approach, Conceptual Misunderstanding, Missing Step, Logical Fallacy, Factual Error, Incomplete Reasoning, Misreading the Problem)'', \\
            ``why\_leads\_to\_wrong\_answer'': ``Explanation of how this error causes wrong answers'' \\
        \} \\
    ], \\
    ``failure\_assignments'': [ \\
        \{ \\
            ``failure\_id'': 1, \\
            ``problem\_idx'': <problem\_idx>, \\
            ``run\_id'': <run\_id>, \\
            ``is\_new\_category'': false, \\
            ``category\_name'': ``Name of existing or new category'', \\
            ``trace\_details'': \{ \\
                ``trace\_specific\_location'': ``Where in the reasoning the error occurred'', \\
                ``trace\_specific\_details'': ``Specific details about what went wrong'' \\
            \} \\
        \} \\
    ] \\
\} \\
\verb|```| \\
\\
Note: ``new\_categories'' should only contain categories that don't exist yet. \\
\end{tcolorbox}

\begin{tcolorbox}[promptbox={Guidance generation prompt}]
\small
\ttfamily
You are an expert at improving language model performance on relation extraction with binary yes/no inference. \\
\\
I have identified the following recurring error categories from model failures. Use them to revise the current instruction prompt so the model can avoid these mistakes. \\
\\
\#\# Current Instruction Prompt \\
\\
\verb|```|text \\
\#CURRENT\_INSTRUCTION\_PROMPT\# \\
\verb|```| \\
\\
\#\# Error Categories \\
\\
\verb|```|text \\
\#ERROR\_CATEGORIES\_TEXT\# \\
\verb|```| \\
\\
\#\# Your Task \\
\\
Generate a revised instruction prompt that: \\
1. Addresses each failure category with specific, actionable advice \\
2. Improves and generalizes binary relation extraction decisions \\
\\
\#\# Critical Constraints \\
\\
- Do not revise the instruction in a way that requires step-by-step reasoning or explanatory output; the task should remain direct binary yes/no inference. \\
- Preserve compatibility with the existing prompt structure where the answer instruction and input template are appended separately. \\
- The goal is ACCURACY, not caution. Never generate guidance that encourages the model to refuse, abstain, or say ``not specified'' when an answer can be provided. \\
\\
\#\# Output Format \\
\\
Return a JSON object with: \\
\{ \\
  ``revised\_instruction\_prompt'': ``the revised instruction prompt text'' \\
\} \\
\end{tcolorbox}

\subsubsection{\rpo{} meta-prompts}
\label{subsec:rpo-meta-prompts}

\begin{tcolorbox}[promptbox={Feedback generation prompt}]
\small
\ttfamily
You are an expert feedback model for a relation extraction inference task. Specifically, you are skilled at providing feedback explaining why a relation extraction system arrived at a particular yes/no decision, for both correct and incorrect predictions. \\
\\
A relation captures the connection between two entities in a sentence by describing their relationship. We will refer to these entities as the subject and object entities. \\
The task requires inferring a binary (yes/no) answer based on whether the query sentence expresses this relation between the subject and the object. \\
\\
You are given an input instance for relation inference below that contains: \\
- A relation and its description \\
- A support instance (an example where the relation holds) \\
- A query sentence \\
- A ground-truth label (yes/no) \\
- A yes/no inference made by another LLM on this instance \\
\\
The ground-truth label indicates whether the LLM inference was correct or incorrect and is provided only as contextual information. The feedback model’s task is to explain the most likely reasoning process that led to the model’s answer, not to re-evaluate, judge, or correct the prediction. In particular: \\
- If the prediction matches the label, explain what cues or evidence likely led to that choice. \\
- If it does not match, explain what misunderstanding, missing evidence, or heuristic likely caused the prediction. \\
\\
Instance: \\
\verb|```| \\
Relation: \#RELATION\# \\
Relation Description: \#RELATION\_DESCRIPTION\# \\
Support Instance: \#SUPPORT\_INSTANCE\# \\
\\
Query: \#QUERY\# \\
\\
Label: \#LABEL\# \\
LLM Inference: \#INFERENCE\# \\
\verb|```| \\
\\
Please reason through the problem, but provide your final feedback only within the <f> and </f> tags. \\
\end{tcolorbox}

\begin{tcolorbox}[promptbox={Mutation prompt}]
\small
\ttfamily
You are an expert prompt generator for a relation extraction inference task. You specialize in revising and improving prompts based on feedback from previous model predictions. \\
\\
A relation captures the connection between two entities in a sentence by describing their relationship. We will refer to these entities as the subject and object entities. \\
The task requires inferring a binary (yes/no) answer based on whether the query sentence expresses this relation between the subject and the object. \\
\\
You are given below a prompt that is used by another LLM to make an inference for the task: \\
\verb|```| \\
\#INFERENCE\_PROMPT\# \\
\verb|```| \\
\\
Using this prompt, another LLM was tested on three instances of the task. Below, you are given the inputs, the inference made by the other LLM, and feedback for each task. \\
\verb|```| \\
Task 1 \\
Relation: \#RELATION\_1\# \\
Relation Description: \#RELATION\_DESCRIPTION\_1\# \\
Support Instance: \#SUPPORT\_INSTANCE\_1\# \\
Query Sentence: \#QUERY\_1\# \\
Ground-Truth Label: \#LABEL\_1\# \\
LLM Inference: \#INFERENCE\_1\# \\
Feedback: \#FEEDBACK\_1\# \\
\\
Task 2 \\
Relation: \#RELATION\_2\# \\
Relation Description: \#RELATION\_DESCRIPTION\_2\# \\
Support Instance: \#SUPPORT\_INSTANCE\_2\# \\
Query Sentence: \#QUERY\_2\# \\
Ground-Truth Label: \#LABEL\_2\# \\
LLM Inference: \#INFERENCE\_2\# \\
Feedback: \#FEEDBACK\_2\# \\
\\
Task 3 \\
Relation: \#RELATION\_3\# \\
Relation Description: \#RELATION\_DESCRIPTION\_3\# \\
Support Instance: \#SUPPORT\_INSTANCE\_3\# \\
Query Sentence: \#QUERY\_3\# \\
Ground-Truth Label: \#LABEL\_3\# \\
LLM Inference: \#INFERENCE\_3\# \\
Feedback: \#FEEDBACK\_3\# \\
\verb|```| \\
\\
Carefully read the inputs, outputs, and feedback to identify problems with the current prompt. \\
Your task is to generate a revised version of the prompt that helps the other LLM generalize better when using it. \\
You may modify, add to, or remove any instructions or content in the current prompt in order to improve the prediction and enhance generalization. \\
\\
Please reason through the problem, but output only the revised prompt enclosed within the <p> and </p> tags. \\
\end{tcolorbox}

\subsubsection{\evoprompt{-DE} meta-prompts}
\label{subsec:evoprompt-meta-prompts}

\begin{tcolorbox}[promptbox={DE operation prompt}]
\small
\ttfamily
Please follow the instruction step-by-step to generate a better prompt. \\
1. Identify the different parts between Prompt 1 and Prompt 2: \\
Prompt 1: Determine whether the named relation holds between the tagged subject and tagged object in the query sentence. Use the support sentence only as an example of the relation. \\
Prompt 2: Decide if the query sentence expresses the requested relation from Subject to Object. Check the relation definition, entity roles, direction, and whether the evidence is explicit or clearly implied. \\
2. Randomly mutate the different parts. \\
3. Combine the different parts with Prompt 3, selectively replace it with the different parts from step 2, and generate a new prompt. \\
Prompt 3: Carefully compare the relation definition with the query sentence, verify that the tagged Subject and Object have the right semantic types and direction, and avoid using irrelevant facts from the support example. \\
4. Crossover the prompt in step 3 with the following basic prompt and generate a final prompt bracketed with \#PROMPT\_OPEN\_TAG\# and \#PROMPT\_CLOSE\_TAG\#: \\
Basic Prompt: You are given a relation name, a relation description, a support sentence that exemplifies the relation, and a query sentence. Answer only yes or no depending on whether the relation holds between the tagged Subject and Object in the query sentence. \\
\\
1. Identifying the different parts between Prompt 1 and Prompt 2: \\
Prompt 1: Determine whether the named relation holds between the tagged subject and tagged object in the query sentence. Use the support sentence only as an example of the relation. \\
Prompt 2: Decide if the query sentence expresses the requested relation from Subject to Object. Check the relation definition, entity roles, direction, and whether the evidence is explicit or clearly implied. \\
Different parts: \\
``named relation holds'' vs ``query sentence expresses the requested relation'' \\
``tagged subject and tagged object'' vs ``Subject to Object'' \\
``Use the support sentence only as an example'' vs ``Check the relation definition, entity roles, direction'' \\
\\
2. Randomly mutate the different parts: \\
``named relation holds'' -> ``specified relation is valid'' \\
``query sentence expresses the requested relation'' -> ``query provides enough evidence for the target relation'' \\
``tagged subject and tagged object'' -> ``marked Subject and Object entities'' \\
``Check the relation definition, entity roles, direction'' -> ``verify the definition, argument roles, semantic types, and direction'' \\
\\
3. Combine the different parts with Prompt 3, selectively replace it with the different parts in step 2 and generate a new prompt: \\
Prompt 3: Carefully compare the relation definition with the query sentence, verify that the tagged Subject and Object have the right semantic types and direction, and avoid using irrelevant facts from the support example. \\
New Prompt: Carefully compare the relation definition with the query sentence and decide whether the specified relation is valid for the marked Subject and Object entities. Verify the definition, argument roles, semantic types, and direction, and use the support sentence only as a guide to the relation type. \\
\\
4. Crossover the prompt in step 3 with the following basic prompt and generate a final prompt bracketed with \#PROMPT\_OPEN\_TAG\# and \#PROMPT\_CLOSE\_TAG\#: \\
Basic Prompt: You are given a relation name, a relation description, a support sentence that exemplifies the relation, and a query sentence. Answer only yes or no depending on whether the relation holds between the tagged Subject and Object in the query sentence. \\
Final Prompt: \#PROMPT\_OPEN\_TAG\#Carefully compare the relation definition with the query sentence and decide whether the specified relation is valid for the tagged Subject and Object entities. Use the support sentence only as an example of the relation type. Verify the argument roles, semantic types, direction of the relation, and whether the query gives explicit or clearly implied evidence. If the relation holds between the Subject and Object in the query sentence, answer ``yes''; otherwise, answer ``no''.\#PROMPT\_CLOSE\_TAG\# \\
\\
Please follow the instruction step-by-step to generate a better prompt. \\
1. Identify the different parts between Prompt 1 and Prompt 2: \\
Prompt 1: <prompt1> \\
Prompt 2: <prompt2> \\
2. Randomly mutate the different parts. \\
3. Combine the different parts with Prompt 3, selectively replace it with the different parts in step 2, and generate a new prompt. \\
Prompt 3: <prompt3> \\
4. Crossover the prompt in step 3 with the following basic prompt and generate a final prompt bracketed with \#PROMPT\_OPEN\_TAG\# and \#PROMPT\_CLOSE\_TAG\#: \\
Basic Prompt: <prompt0> \\
\\
1. \\
\end{tcolorbox}

\subsection{Optimized prompts by \qwensmall{} on \tacred}
\label{subsec:optimized-prompts}

\subsubsection{\rpo{}-optimized prompt at iteration 5 and its second-stage refinements}
\label{subsec:rpo-5th}

\begin{tcolorbox}[promptbox={\rpo{}-optimized prompt}]
\small
\ttfamily
You are given a relation name, a description of the relation in brackets, a support sentence exemplifying the relation, and a query sentence. A relation connects the Subject and Object entities, which are indicated with the subject and object tags, respectively. \\
\\
To determine if the relation holds between the Subject and Object entities in the query sentence, strictly follow these steps: \\
1. **Verify the subject and object types**: \\
   - For relations starting with "org:", the **subject must be an organization** (not a person or role). \\
   - The **object must align with the relation's definition** (e.g., "org:subsidiaries" requires a subsidiary entity, not a role or category; "org:political/religious\_affiliation" accepts a religious or political category, not a person). \\
2. **Check the relationship**: Confirm that the query sentence explicitly links the Subject and Object as per the relation's description. For example, "org:subsidiaries" requires a hierarchical connection between an organization and its subsidiary (e.g., "Company A owns Company B"), not a personal or abstract relationship. \\
3. **Recognize object types**: \\
   - The object can be a **specific entity** (e.g., "US-based Ryder Public Transportation Services") or a **category** (e.g., "Christian", "Islamist") if the relation explicitly allows it. \\
   - Avoid treating abstract terms (e.g., "Christian") as roles or people unless the relation's description explicitly permits it. \\
4. **Focus on explicit connections**: Only answer "yes" if the query sentence directly establishes the relation between the subject and object as defined. For example, "org:number\_of\_employees/members" requires a numerical value (e.g., "42,000") to indicate the count, not a vague term like "many." \\
\\
If the relation holds, answer "yes"; otherwise, answer "no". \\
\end{tcolorbox}

\begin{tcolorbox}[promptbox={\lpo{}-optimized prompt}]
\small
\ttfamily
You are given a relation name, a description of the relation in brackets, a support sentence exemplifying the relation, and a query sentence. A relation connects the Subject and Object entities, which are indicated with the subject and object tags, respectively. \\
\\
To determine if the relation holds between the Subject and Object entities in the query sentence, strictly follow these steps: \\
1. **Verify the subject and object types**: \\
   - For relations starting with "org:", the **subject must be an organization** (not a person or role). \\
   - For per: relations, the subject must be a person. \\
   - The **object must align with the relation's definition** (e.g., "org:subsidiaries" requires a subsidiary entity, not a role or category; "org:political/religious\_affiliation" accepts a religious or political category, not a person). \\
2. **Check the relationship**: Confirm that the query sentence explicitly states the link between the Subject and Object as per the relation's description, using direct assertions (e.g., "is from," "was born in," "works for") rather than indirect or implied connections. \\
3. **Recognize object types**: \\
   - The object can be a **specific entity** (e.g., "US-based Ryder Public Transportation Services") or a **explicitly permitted category (e.g., 'Christian,' 'Islamist,' 'Lakota') if the relation explicitly allows it, but avoid treating nationalities, tribes, or administrative regions as valid objects unless explicitly permitted.** \\
   - Avoid treating abstract terms (e.g., "Christian") as roles or people unless the relation's description explicitly permits it. \\
4. **Focus on explicit connections**: Only answer "yes" if the query sentence directly establishes the relation between the subject and object as defined. For example, "org:number\_of\_employees/members" requires a numerical value (e.g., "42,000") to indicate the count, not a vague term like "many." \\
\\
If the relation holds, answer "yes"; otherwise, answer "no". \\
\end{tcolorbox}

\begin{tcolorbox}[promptbox={\greater{}-optimized prompt}]
\small
\ttfamily
I are given a relation name, a description of the relation in brackets, a support sentence exemplifying the relation, and a query sentence. A relation connects the Subject and Object entities, which are indicated with the subject and object tags, respectively. \\
\\
To determine if the relation holds between the Subject and Object entities in the query sentence, strictly follow these steps: \\
1. **Verify the subject and object types**: \\
   - For relations starting with "org:", the **subject must be an organization** (not a person or role). \\
   - The **object must align with the relation's definition** (e.g., "org:subsidiaries" requires a subsidiary entity, not a role or category; "org:political/religious\_affiliation" accepts a religious or political category, not a person). \\
2. **Check the relationship**: Confirm that the query sentence explicitly links the Subject and Object as per the relation's description. For example, "org:subsidiaries" requires a hierarchical connection between an organization and its subsidiary (e.g., "Company A owns Company B"), not a personal or abstract relationship. \\
3. **Recognize object types**: \\
   - The object can be a **specific entity** (e.g., "US-based Ryder Public Transportation Services") or a **category** (e.g., "Christian", "Islamist") if the relation explicitly allows it. \\
   - Avoid treating abstract terms (e.g., "Christian") as roles or people unless the relation's description explicitly permits it. \\
4. **Focus on explicit connections**: Only answer "yes" if the query sentence directly establishes the relation between the subject and object as defined. For example, "org:number\_of\_employees/members" requires a numerical value (e.g., "42,000") to indicate the count, not a vague term like "many." \\
\\
If the relation holds, answer "yes"; otherwise, answer "no". \\
\end{tcolorbox}

\begin{tcolorbox}[promptbox={\greater{-TG}-optimized prompt}]
\small
\ttfamily
You are given a relation name, a description of the relation in brackets, a support sentence exemplifying the relation, and a query sentence. A relation connects the Subject and Object entities, which are indicated with the subject and object tags, respectively. \\
\\
To determine if the relation holds between the Subject and Object entities in the query sentence, strictly follow these steps: \\
1. **Verify the subject and object types**: \\
   - For relations starting with "org:", the **subject must be an organization** (not a person or role). \\
   - The **object must align with the relation's definition** (e.g., "org:subsidiaries" requires a subsidiary entity, not a role or category; "org:political/religious\_affiliation" accepts a religious or political category, not a person). \\
2. **Check the relationship**: Confirm that the query sentence explicitly links the Subject and Object as per the relation's description. For example, "org:subsidiaries" requires a hierarchical connection between an organization and its subsidiary (e.g., "Company A owns Company B"), not a personal or abstract relationship. \\
3. **Recognize object types**: \\
   - The object can be a **specific entity** (e.g., "US-based Ryder Public Transportation Services") or a **category** (e.g., "Christian", "Islamist") if the relation explicitly allows it. \\
   - Avoid treating abstract terms (e.g., "Christian") as roles or people unless the relation's description explicitly permits it. \\
4. **Focus on explicit connections**: Only answer "yes" if the query sentence directly establishes the relation between the subject and object as defined. For example, "org:number\_of\_employees/members" requires a numerical value (e.g., "42,000") to indicate the count, not a vague term like "many." \\
\\
If the relation holds, answer "your"; otherwise, answer "no". \\
\end{tcolorbox}

\begin{tcolorbox}[promptbox={\gradpo{}-optimized prompt}]
\small
\ttfamily
You are given a relation name, a definition of the relation in bracketed text, a support example sentence exemplifying the relation, and a query inquiry sentence. A relation connects the Subject and Object entities, which are indicated with the subject and object tags, respectively. \\
\\
To determine if the relation holds between the Subject and Object entities in the query sentence, strictly follow these steps: \\
1. **Verify the subject and object types**: \\
   - For relations starting with "org:", the **subject must be an organization** (not a person or role). \\
   - The **object must align with the relation's definition** (e.g., "org:subsidiaries" requires a subsidiary entity, not a role or category; "org:political/religious\_affiliation" accepts a religious or political category, not a person). \\
2. **Check the relationship**: Confirm that the query sentence explicitly links the Subject and Object as per the relation's description. For example, "org:subsidiaries" requires a hierarchical connection between an organization and its subsidiary (e.g., "Company A owns Company B"), not a personal or abstract relationship. \\
3. **Recognize object types**: \\
   - The object can be a **specific entity** (e.g., "US-based Ryder Public Transportation Services") or a **category** (e.g., "Christian", "Islamist") if the relation explicitly allows it. \\
   - Avoid treating abstract terms (e.g., "Christian") as roles or people unless the relation's definition explicitly permits it. \\
4. **Focus on explicit connections**: Only answer "yes" if the query sentence directly establishes the relation between the subject and object as defined. For example, "org:number\_of\_employees/members" requires a numerical value (e.g., "42,000") to indicate the count, not a vague term like "many." \\
\\
If the relation holds, answer "yes"; otherwise, answer "no". \\
\end{tcolorbox}

\begin{tcolorbox}[promptbox={\gradpoprob{}-optimized prompt}]
\small
\ttfamily
You are given a relation name, a description of the relation in [brackets], a support sentence exemplifying the relation, and a query sentence. A relation connects the Subject and Object entities, which are indicated with the subject and object tags, respectively. \\
\\
To determine if the relation holds between the Subject and Object entities in the query sentence, strictly follow these steps: \\
1. **Verify the subject and object types**: \\
   - For relations starting with "org:", the subject must be an organization (not a person or role). \\
   - The object must align with the relation's definition (e.g., "org:subsidiaries" requires a subsidiary entity, not a role or category; "org:political/religious\_affiliation" accepts a religious or political category, not a person). \\
2. **Check the relationship**: Confirm that the query sentence explicitly links the Subject and Object as per the relation's description. For example, "org:subsidiaries" requires a hierarchical connection between an organization and its subsidiary (e.g., "Company A owns Company B"), not a personal or abstract relationship. \\
3. **Recognize object types**: \\
   - The object can be a specific entity (e.g., "US-based Ryder Public Transportation Services") or a category (e.g., "Christian", "Islamist") if the relation explicitly allows it. \\
   - Avoid treating abstract terms (e.g., "Christian") as roles or people unless the relation's description explicitly permits it. \\
4. **Focus on explicit connections**: Only answer "yes" if the query sentence directly establishes the relation between the subject and object as defined. For example, "org:number\_of\_employees/members" requires a numerical value (e.g., "42,000") to indicate the count, not a vague term like "many." \\
\\
If the relation holds, answer "yes"; otherwise, answer "no". \\
\end{tcolorbox}

\subsubsection{\rpo{}-optimized prompt at iteration 10 and its second-stage refinements}
\label{subsec:rpo-10th}

\begin{tcolorbox}[promptbox={\rpo{}-optimized prompt}]
\small
\ttfamily
You are given a relation name, a description of the relation in brackets, a support sentence exemplifying the relation, and a query sentence. A relation connects the Subject and Object entities, which are indicated with the subject and object tags, respectively. \\
\\
To determine if the relation holds between the Subject and Object entities in the query sentence, strictly follow these steps: \\
1. **Verify the subject and object types**: The subject must be the entity type expected by the relation (e.g., "org" for organization, "per" for person). The object must match the relation's definition (e.g., "org:stateorprovince\_of\_headquarters" requires the object to be a state/province, not a city or location). If the subject is a location (e.g., "Kennedy Space Center"), it must be explicitly tied to an organization (e.g., NASA) as its headquarters. \\
2. **Check the relationship**: Confirm that the query sentence explicitly links the Subject and Object as per the relation's description. For example, "org:stateorprovince\_of\_headquarters" requires the subject to be an organization (e.g., "NASA") and the object to be a state/province (e.g., "Florida"), not a location (e.g., "Kennedy Space Center"). For "org:political/religious\_affiliation," the object must be the organization's actual political or religious affiliation (e.g., "Islamist"), not a role, descriptor, or faction. \\
3. **Avoid role vs. person confusion**: If the object is a role (e.g., "chief executive") or title (e.g., "Rep."), it does not satisfy relations requiring a person (e.g., "employee\_of"). Similarly, ensure the subject is the correct entity type (e.g., "org" for organizations, "per" for people). For "org:stateorprovince\_of\_headquarters," the subject must be an organization (e.g., "NASA"), not a location (e.g., "Kennedy Space Center"), unless the sentence explicitly states the location is the organization's headquarters. \\
4. **Focus on explicit connections**: Only answer "yes" if the query sentence directly establishes the relation between the subject and object as defined. This includes ensuring the object is the correct entity type (e.g., "state/province" for headquarters, "political/religious affiliation" for org:political/religious\_affiliation) and that the subject is the correct entity type for the relation. \\
\\
If the relation holds, answer "yes"; otherwise, answer "no". \\
\end{tcolorbox}

\begin{tcolorbox}[promptbox={\lpo{}-optimized prompt}]
\small
\ttfamily
You are given a relation name, a description of the relation in brackets, a support sentence exemplifying the relation, and a query sentence. A relation connects the Subject and Object entities, which are indicated with the subject and object tags, respectively. \\
\\
To determine if the relation holds between the Subject and Object entities in the query sentence, strictly follow these steps: \\
1. **Verify the subject and object types**: The subject must be the entity type expected by the relation (e.g., "org" for organization, "per" for person). The object must match the relation's definition (e.g., "org:stateorprovince\_of\_headquarters" requires the object to be a state/province, not a city or location). must be a state/province and not a city. If the subject is a location (e.g., "Kennedy Space Center"), it must be explicitly tied to an organization (e.g., NASA) as its headquarters. \\
2. **Check the relationship**: Confirm that the query sentence explicitly links the Subject and Object as per the relation's description. Ensure the query sentence explicitly states the relation between the subject and object as per the relation's description (e.g., "state/province" for headquarters, "family member" for other\_family). For example, "org:stateorprovince\_of\_headquarters" requires the subject to be an organization (e.g., "NASA") and the object to be a state/province (e.g., "Florida"), not a location (e.g., "Kennedy Space Center"). For "org:political/religious\_affiliation," the object must be the organization's actual political or religious affiliation (e.g., "Islamist"), not a role, descriptor, or faction. \\
3. **Avoid role vs. person confusion**: If the object is a role (e.g., "chief executive") or title (e.g., "Rep."), it does not satisfy relations requiring a person (e.g., "employee\_of"). Possessive pronouns (e.g., "his", "her") do not satisfy relations requiring a person (e.g., "other\_family"). Similarly, ensure the subject is the correct entity type (e.g., "org" for organizations, "per" for people). For "org:stateorprovince\_of\_headquarters," the subject must be an organization (e.g., "NASA"), not a location (e.g., "Kennedy Space Center"), unless the sentence explicitly states the location is the organization's headquarters. \\
4. **Focus on explicit connections**: Only answer "yes" if the query sentence directly establishes the relation between the subject and object as defined. This includes ensuring the object is the correct entity type (e.g., "state/province" for headquarters, "political/religious affiliation" for org:political/religious\_affiliation) and that the subject is the correct entity type for the relation. \\
\\
If the relation holds, answer "yes"; otherwise, answer "no". \\
\end{tcolorbox}

\begin{tcolorbox}[promptbox={\greater{}-optimized prompt}]
\small
\ttfamily
Class are given a relation name, a description of the relation in brackets, a support sentence exemplifying the relation, and a query sentence. A relation connects the Subject and Object entities, which are indicated with the subject and object tags, respectively. \\
\\
To determine if the relation holds between the Subject and Object entities in the query sentence, strictly follow these steps: \\
1. **Verify the subject and object types**: The subject must be the entity type expected by the relation (e.g., "org" for organization, "per" for person). The object must match the relation's definition (e.g., "org:stateorprovince\_of\_headquarters" requires the object to be a state/province, not a city or location). If the subject is a location (e.g., "Kennedy Space Center"), it must be explicitly tied to an organization (e.g., NASA) as its headquarters. \\
2. **Check the relationship**: Confirm that the query sentence explicitly links the Subject and Object as per the relation's description. For example, "org:stateorprovince\_of\_headquarters" requires the subject to be an organization (e.g., "NASA") and the object to be a state/province (e.g., "Florida"), not a location (e.g., "Kennedy Space Center"). For "org:political/religious\_affiliation," the object must be the organization's actual political or religious affiliation (e.g., "Islamist"), not a role, descriptor, or faction. \\
3. **Avoid role vs. person confusion**: If the object is a role (e.g., "chief executive") or title (e.g., "Rep."), it does not satisfy relations requiring a person (e.g., "employee\_of"). Similarly, ensure the subject is the correct entity type (e.g., "org" for organizations, "per" for people). For "org:stateorprovince\_of\_headquarters," the subject must be an organization (e.g., "NASA"), not a location (e.g., "Kennedy Space Center"), unless the sentence explicitly states the location is the organization's headquarters. \\
4. **Focus on explicit connections**: Only answer "yes" if the query sentence directly establishes the relation between the subject and object as defined. This includes ensuring the object is the correct entity type (e.g., "state/province" for headquarters, "political/religious affiliation" for org:political/religious\_affiliation) and that the subject is the correct entity type for the relation. \\
\\
If the relation holds, answer "yes"; otherwise, answer "no". \\
\end{tcolorbox}

\begin{tcolorbox}[promptbox={\greater{-TG}-optimized prompt}]
\small
\ttfamily
You are given a relation name, a description of the relation in brackets, a support sentence exemplifying the relation, and a query sentence. A relation connects the Subject and Object entities, which are indicated with the subject and object tags, respectively. \\
\\
To determine if the relation holds between the Subject and Object entities in the query sentence, strictly follow these steps: \\
1. **Verify the subject and object types**: The subject must be the entity type expected by the relation (e.g., "org" for organization, "per" for person). The object must match the relation's definition (e.g., "org:stateorprovince\_of\_headquarters" requires the object to be a state/province, not a city or location). If the subject is a location (e.g., "Kennedy Space Center"), it must be explicitly tied to an organization (e.g., NASA) as its headquarters. \\
2. **Check the relationship**: Confirm that the query sentence explicitly links the Subject and Object as per the relation's description. For example, "org:stateorprovince\_of\_headquarters" requires the subject to be an organization (e.g., "NASA") and the object to be a state/province (e.g., "Florida"), not a location (e.g., "Kennedy Space Center"). For "org:political/religious\_affiliation," the object must be the organization's actual political or religious affiliation (e.g., "Islamist"), not a role, descriptor, or faction. \\
3. **Avoid role vs. person confusion**: If the object is a role (e.g., "chief executive") or title (e.g., "Rep."), it does not satisfy relations requiring a person (e.g., "employee\_of"). Similarly, ensure the subject is the correct entity type (e.g., "org" for organizations, "per" for people). For "org:stateorprovince\_of\_headquarters," the subject must be an organization (e.g., "NASA"), not a location (e.g., "Kennedy Space Center"), unless the sentence explicitly states the location is the organization's headquarters. \\
4. **Focus on explicit connections**: Only answer "yes" if the query sentence directly establishes the relation between the subject and object as defined. This includes ensuring the object is the correct entity type (e.g., "state/province" for headquarters, "political/religious affiliation" for org:political/religious\_affiliation) and that the subject is the correct entity type for the relation. \\
\\
If the relation holds, answer "1"; otherwise, answer "no". \\
\end{tcolorbox}

\begin{tcolorbox}[promptbox={\gradpo{}-optimized prompt}]
\small
\ttfamily
You are given a relation property, a description of the relation in curly braces, a support sentence exemplifying the relation, and a query sentence. A relation connects the Subject and Object entities, which are indicated with the subject and object tags, respectively. \\
\\
To determine if the relation holds between the Subject and Object entities in the query sentence, strictly follow these steps: \\
1. **Verify the subject and object types**: The subject must be the entity type expected by the relation (e.g., "org" for organization, "per" for person). The object must match the relation's definition (e.g., "org:stateorarea\_of\_headquarters" requires the object to be an area, not a city or location). If the subject is a location (e.g., "Kennedy Space Center"), it must be explicitly tied to an organization (e.g., NASA) as its headquarters. \\
2. **Check the relationship**: Confirm that the query sentence explicitly links the Subject and Object as per the relation's description. For example, "org:stateorarea\_of\_headquarters" requires the subject to be an organization (e.g., "NASA") and the object to be an area (e.g., "Florida"), not a location (e.g., "Kennedy Space Center"). For "org:political/religious\_affiliation," the object must be the organization's actual political or religious affiliation (e.g., "Islamist"), not a role, descriptor, or faction. \\
3. **Avoid role vs. person confusion**: If the object is a role (e.g., "chief executive") or title (e.g., "Rep."), it does not satisfy relations requiring a person (e.g., "employee\_of"). Similarly, ensure the subject is the correct entity type (e.g., "org" for organizations, "per" for people). For "org:stateorarea\_of\_headquarters," the subject must be an organization (e.g., "NASA"), not a location (e.g., "Kennedy Space Center"), unless the sentence explicitly states the location is the organization's headquarters. \\
4. **Focus on explicit connections**: Only confirm if the query sentence directly establishes the relation between the subject and object as defined. This includes ensuring the object is the correct entity type (e.g., "area" for headquarters, "political/religious affiliation" for org:political/religious\_affiliation) and that the subject is the correct entity type for the relation. \\
\\
If the relation holds, answer "yes"; otherwise, answer "no". \\
\end{tcolorbox}

\begin{tcolorbox}[promptbox={\gradpoprob{}-optimized prompt}]
\small
\ttfamily
You are given a relation , a description of the relation in brackets, a support sentence exemplifying the relation, and a query sentence. A relation connects the Subject and Object entities, which are indicated with the subject and object tags, respectively. \\
\\
To determine if the relation holds between the Subject and Object entities in the query sentence, strictly follow these steps: \\
1. **Verify the subject and object types**: The subject must be the entity type expected by the relation (e.g., "org" for organization, "per" for person). The object must match the relation's definition (e.g., "org:stateorprovince\_of\_headquarters" requires the object to be a state/province, not a city or location). If the subject is a location (e.g., "Kennedy Space Center"), it must be explicitly tied to an organization (e.g., NASA) as its headquarters. \\
2. **Check the relationship**: Confirm that the query sentence explicitly links the Subject and Object as per the relation's description. For example, "org:stateorprovince\_of\_headquarters" requires the subject to be an organization (e.g., "NASA") and the object to be a state/province (e.g., "Florida"), not a location (e.g., "Kennedy Space Center"). For "org:political/religious\_affiliation," the object must be the organization's actual political or religious affiliation (e.g., "Islamist"), not a role, descriptor, or faction. \\
3. **Avoid role vs. person confusion**: If the object is a role (e.g., "chief executive") or title (e.g., "Rep."), it does not satisfy relations requiring a person (e.g., "employee\_of"). Similarly, ensure the subject is the correct entity type (e.g., "org" for organizations, "per" for people). For "org:stateorprovince\_of\_headquarters," the subject must be an organization (e.g., "NASA"), not a location (e.g., "Kennedy Space Center"), unless the sentence explicitly states the location is the organization's headquarters. \\
4. **Focus on explicit connections**: Only answer "yes" if the query sentence directly establishes the relation between the subject and object as defined. This includes ensuring the object is the correct entity type (e.g., "state/province" for headquarters, "political/religious affiliation" for org:political/religious\_affiliation) and that the subject is the correct entity type for the relation. \\
\\
If the relation holds, answer "yes"; otherwise, answer "no". \\
\end{tcolorbox}

\subsubsection{\evoprompt{-DE}-optimized prompt at iteration 5 and its second-stage refinements}
\label{subsec:evoprompt-5th}

\begin{tcolorbox}[promptbox={\evoprompt{-DE}-optimized prompt}]
\small
\ttfamily
You are given a relation name, its description, a support sentence that exemplifies the relation, and a query sentence. Analyze the relation description and the query sentence. The Subject and Object are labeled in the query sentence. Determine if the relation is expressed between the marked entities. Check the relation definition, the semantic type of each entity, the direction from Subject to Object, and whether the evidence is explicit or strongly implied. Treat the support sentence as a guide to the relation, not evidence for the query. Confirm that the relation direction is Subject-to-Object, that the Object has the required type or value, and that the query provides sufficient explicit or contextually implied evidence. Ignore any irrelevant entities or facts not directly stated in the query. If the relation holds in the query sentence, answer "yes"; otherwise answer "no". \\
\end{tcolorbox}

\begin{tcolorbox}[promptbox={\greater{}-optimized prompt}]
\small
\ttfamily
I are given a relation name, its description, a support sentence that exemplifies the relation, and a query sentence. Analyze the relation description and the query sentence. The Subject and Object are labeled in the query sentence. Determine if the relation is expressed between the marked entities. Check the relation definition, the semantic type of each entity, the direction from Subject to Object, and whether the evidence is explicit or strongly implied. Treat the support sentence as a guide to the relation, not evidence for the query. Confirm that the relation direction is Subject-to-Object, that the Object has the required type or value, and that the query provides sufficient explicit or contextually implied evidence. Ignore any irrelevant entities or facts not directly stated in the query. If the relation holds in the query sentence, answer "yes"; otherwise answer "no". \\
\end{tcolorbox}

\begin{tcolorbox}[promptbox={\greater{-TG}-optimized prompt}]
\small
\ttfamily
You are given a relation name, its description, a support sentence that exemplifies the relation, and a query sentence. Analyze the relation description and the query sentence. The Subject and Object are labeled in the query sentence. Determine if the relation is expressed between the marked entities. Check the relation definition, the semantic type of each entity, the direction from Subject to Object, and whether the evidence is explicit or strongly implied. Treat the support sentence as a guide to the relation, not evidence for the query. Confirm that the relation direction is Subject-to-Object, that the Object has the required type or value, and that the query provides sufficient explicit or contextually implied evidence. Ignore any irrelevant entities or facts not directly stated in the query. If the relation holds in the query sentence, answer "your"; otherwise answer "no". \\
\end{tcolorbox}

\begin{tcolorbox}[promptbox={\gradpogen{}-optimized prompt}]
\small
\ttfamily
You are told given a relation name, its description, a support text that demonstrates the relation, and a query sentence. Analyze the relation description and the query sentence. The Subject and Object are labeled in the query sentence. Evaluate if the relation is expressed between the marked entities. Check the relation definition, the semantic type of each entity, the direction from Subject to Object, and whether the evidence is explicit or strongly implied. Treat the support sentence as a guide to the relation, not evidence for the query. Confirm that the relation direction is Subject-to-Object, that the Object has the required type or value, and that the query provides sufficient explicit or contextually implied evidence. Ignore any irrelevant entities or facts not directly stated in the query. If the relation holds in the query sentence, answer "yes"; otherwise answer "no". \\
\end{tcolorbox}

\begin{tcolorbox}[promptbox={\gradpoprob{}-optimized prompt}]
\small
\ttfamily
The provided relation name, its description, a support example that exemplifies the relation, and a query sentence. Analyze the relation description and the query sentence. The Subject and Object are labeled in the query sentence. Check if the relation is expressed between the marked entities. Check the relation definition, the semantic type of each entity, the direction from Subject to Object, and whether the evidence is explicit or strongly implied. Treat the support sentence as a guide to the relation, not evidence for the query. Confirm that the relation direction is Subject-to-Object, that the Object has the required type or value, and that the query provides sufficient explicit or contextually implied evidence. Ignore any irrelevant entities or facts not directly stated in the query. If the relation holds in the query sentence, answer "yes"; otherwise answer "no". \\
\end{tcolorbox}

\subsubsection{\evoprompt{-DE}-optimized prompt at iteration 10 and its second-stage refinements}
\label{subsec:evoprompt-10th}

\begin{tcolorbox}[promptbox={\evoprompt{-DE}-optimized prompt}]
\small
\ttfamily
You are given a relation name, its description, an example sentence, and a query. Analyze the relation and query for consistency and check whether the relation is properly instantiated between the labeled Subject and Object. Analyze the relation definition, semantic roles and types of entities, and the direction of the relation. Determine whether the evidence is directly mentioned or clearly implied. Use the support sentence only as a template for the relation. Do not consider external or unrelated information. Ensure the relation direction is from Subject to Object, that the entities match the required semantic roles, and that the evidence is clearly or strongly expressed. Answer 'yes' only if the relation is clearly or strongly expressed between the Subject and Object in the query; otherwise, answer 'no'. \\
\end{tcolorbox}

\begin{tcolorbox}[promptbox={\lpo{}-optimized prompt}]
\small
\ttfamily
You are given a relation name, its description, an example sentence, and a query. Analyze the relation and query for consistency and check whether the relation is clearly instantiated between the labeled Subject and Object. Analyze the relation definition, semantic roles and types of entities, and the direction of the relation. Determine whether the evidence is strongly mentioned or clearly implied. Use the support sentence only as a template for the relation. Do not consider external or unrelated information. Ensure the relation direction is strictly from Subject to Object, that the entities correctly match the required semantic roles, and that the evidence is strongly mentioned or strongly expressed. Answer 'yes' only if the relation is explicitly mentioned or strongly expressed between the Subject and Object in the query; otherwise, answer 'no'. \\
\end{tcolorbox}

\begin{tcolorbox}[promptbox={\greater{}-optimized prompt}]
\small
\ttfamily
Given are given a relation name, its description, an example sentence, and a query. Analyze the relation and query for consistency and check whether the relation is properly instantiated between the labeled Subject and Object. Analyze the relation definition, semantic roles and types of entities, and the direction of the relation. Determine whether the evidence is directly mentioned or clearly implied. Use the support sentence only as a template for the relation. Do not consider external or unrelated information. Ensure the relation direction is from Subject to Object, that the entities match the required semantic roles, and that the evidence is clearly or strongly expressed. Answer 'yes' only if the relation is clearly or strongly expressed between the Subject and Object in the query; otherwise, answer 'no'. \\
\end{tcolorbox}

\begin{tcolorbox}[promptbox={\gradpo{}-optimized prompt}]
\small
\ttfamily
You are given a relation name, its description, an example sentence, and a query. Analyze the relation and query for consistency and check whether the relation is properly instantiated between the labeled Subject and Object. Analyze the relation definition, semantic roles and types of entities, and the direction of the relation. Determine whether the evidence is directly mentioned or clearly implied. Use the provided sentence only as a template for the relation. Do not consider external or unrelated information. Ensure the relation direction is from Subject to Object, that the entities match the required semantic roles, and that the evidence is clearly or strongly expressed. Respond 'yes' only if the relation is clearly or strongly expressed between the Subject and Object in the query; otherwise, respond 'no'. \\
\end{tcolorbox}

\subsubsection{\etgpo{}-optimized prompt and its second-stage refinements}
\label{subsec:etgpo-optimized-prompts}

\begin{tcolorbox}[promptbox={\etgpo{}-optimized prompt}]
\small
\ttfamily
You are given a relation name, a description of the relation in brackets, a support sentence exemplifying the relation, and a query sentence. A relation connects the Subject and the Object entities indicated by subject/object tags. You must determine if the relation holds between the tagged entities in the query sentence, following these rules: \\
\\
1. **Entity Resolution**: Treat abbreviations (e.g., 'Minn.') as their full forms (e.g., 'Minnesota'), and ensure the object matches the required entity type (e.g., state/province vs. city vs. country). \\
2. **Explicit Matching**: Only use information explicitly stated in the query. Do not infer events (e.g., death) or relationships (e.g., familial ties) unless directly mentioned. \\
3. **Relation Type Precision**: Match the exact relationship type specified (e.g., 'familial' vs. 'friendship'). Do not conflate non-familial roles with familial relations. \\
4. **Role Implication Awareness**: Recognize that certain roles (e.g., 'managing principal') inherently imply specific relationships (e.g., 'shareholder') even if not explicitly stated. \\
5. **Geographic Hierarchy**: Distinguish between sovereign countries (e.g., 'Cuba') and subnational entities (e.g., 'California'). A country cannot satisfy a relation requiring a state/province. \\
\\
Answer 'yes' if the relation holds under these constraints; otherwise, answer 'no'. \\
\end{tcolorbox}

\begin{tcolorbox}[promptbox={\lpo{}-optimized prompt}]
\small
\ttfamily
You are given a relation name, a description of the relation in brackets, a support sentence exemplifying the relation, and a query sentence. A relation connects the Subject and the Object entities indicated by subject/object tags. You must determine if the relation holds between the tagged entities in the query sentence, following these rules: \\
\\
1. **Entity Resolution**: Treat abbreviations (e.g., 'Minn.') as their full forms (e.g., 'Minnesota'), and ensure the object matches the required entity type (e.g., state/province vs. city vs. country, URLs). \\
2. **Explicit Matching**: Only use information explicitly stated in the query. Do not infer events (e.g., death) or relationships (e.g., familial ties) unless directly mentioned. \\
3. **Relation Type Precision**: Match the exact relationship type specified (e.g., 'familial' vs. 'friendship'). Do not conflate non-familial roles with familial relations. Organizational components such as congresses may imply membership if explicitly named. \\
4. **Role Implication Awareness**: Recognize that certain roles (e.g., 'managing principal') inherently imply specific relationships (e.g., 'shareholder') even if not explicitly stated. Include organizational roles (e.g., 'member', 'committee') that directly indicate membership. \\
5. **Geographic Hierarchy**: Distinguish between sovereign countries (e.g., 'Cuba') and subnational entities (e.g., 'California'). A country cannot satisfy a relation requiring a state/province. Subnational entities such as states or provinces are not sovereign. \\
\\
Answer 'yes' if the relation holds under these constraints; otherwise, answer 'no'. \\
\end{tcolorbox}

\begin{tcolorbox}[promptbox={\greater{}-optimized prompt}]
\small
\ttfamily
Given are given a relation name, a description of the relation in brackets, a support sentence exemplifying the relation, and a query sentence. A relation connects the Subject and the Object entities indicated by subject/object tags. You must determine if the relation holds between the tagged entities in the query sentence, following these rules: \\
\\
1. **Entity Resolution**: Treat abbreviations (e.g., 'Minn.') as their full forms (e.g., 'Minnesota'), and ensure the object matches the required entity type (e.g., state/province vs. city vs. country). \\
2. **Explicit Matching**: Only use information explicitly stated in the query. Do not infer events (e.g., death) or relationships (e.g., familial ties) unless directly mentioned. \\
3. **Relation Type Precision**: Match the exact relationship type specified (e.g., 'familial' vs. 'friendship'). Do not conflate non-familial roles with familial relations. \\
4. **Role Implication Awareness**: Recognize that certain roles (e.g., 'managing principal') inherently imply specific relationships (e.g., 'shareholder') even if not explicitly stated. \\
5. **Geographic Hierarchy**: Distinguish between sovereign countries (e.g., 'Cuba') and subnational entities (e.g., 'California'). A country cannot satisfy a relation requiring a state/province. \\
\\
Answer 'yes' if the relation holds under these constraints; otherwise, answer 'no'. \\
\end{tcolorbox}

\begin{tcolorbox}[promptbox={\greater{-TG}-optimized prompt}]
\small
\ttfamily
You are given a relation name, a description of the relation in brackets, a support sentence exemplifying the relation, and a query sentence. A relation connects the Subject and the Object entities indicated by subject/object tags. You must determine if the relation holds between the tagged entities in the query sentence, following these rules: \\
\\
1. **Entity Resolution**: Treat abbreviations (e.g., 'Minn.') as their full forms (e.g., 'Minnesota'), and ensure the object matches the required entity type (e.g., state/province vs. city vs. country). \\
2. **Explicit Matching**: Only use information explicitly stated in the query. Do not infer events (e.g., death) or relationships (e.g., familial ties) unless directly mentioned. \\
3. **Relation Type Precision**: Match the exact relationship type specified (e.g., 'familial' vs. 'friendship'). Do not conflate non-familial roles with familial relations. \\
4. **Role Implication Awareness**: Recognize that certain roles (e.g., 'managing principal') inherently imply specific relationships (e.g., 'shareholder') even if not explicitly stated. \\
5. **Geographic Hierarchy**: Distinguish between sovereign countries (e.g., 'Cuba') and subnational entities (e.g., 'California'). A country cannot satisfy a relation requiring a state/province. \\
\\
Answer 'none' if the relation holds under these constraints; otherwise, answer 'no'. \\
\end{tcolorbox}

\begin{tcolorbox}[promptbox={\gradpo{}-optimized prompt}]
\small
\ttfamily
You are supplied a relation name, a description of the relation in curly brackets, a support sentence exemplifying the relation, and a query sentence. A relation connects the Subject and the Object entities indicated by subject/object tags. You must determine if the relation holds between the tagged entities in the query sentence, following these rules: \\
\\
1. **Entity Resolution**: Treat abbreviations (e.g., 'Minn.') as their full forms (e.g., 'Minnesota'), and ensure the object matches the required entity type (e.g., state/province vs. city vs. country). \\
2. **Explicit Matching**: Only use information explicitly stated in the query. Do not infer events (e.g., death) or relationships (e.g., familial ties) unless directly mentioned. \\
3. **Relation Type Precision**: Match the exact relationship type specified (e.g., 'familial' vs. 'friendship'). Do not conflate non-familial roles with familial relations. \\
4. **Role Implication Awareness**: Recognize that certain roles (e.g., 'managing principal') inherently imply specific relationships (e.g., 'shareholder') even if not explicitly stated. \\
5. **Geographic Hierarchy**: Distinguish between sovereign countries (e.g., 'Cuba') and subnational entities (e.g., 'California'). A country cannot satisfy a relation requiring a state/province. \\
\\
Answer 'yes' if the relation holds under these constraints; otherwise, answer 'no'. \\
\end{tcolorbox}

\begin{tcolorbox}[promptbox={\gradpoprob{}-optimized prompt}]
\small
\ttfamily
Answer the question by determining if the relation holds under these constraints. A relation connects the Subject and the Object entities indicated by subject/object tags. You must determine if the relation holds between the tagged entities in the query sentence, following these rules: \\
\\
1. **Entity Resolution**: Treat abbreviations (e.g., 'Minn.') as their full forms (e.g., 'Minnesota'), and ensure the object matches the required entity type (e.g., state/province vs. city vs. country). \\
2. **Explicit Matching**: Only use information explicitly stated in the query. Do not infer events (e.g., death) or relationships (e.g., familial ties) unless directly mentioned. \\
3. **Relation Type Precision**: Match the exact relationship type specified (e.g., 'familial' vs. 'friendship'). Do not conflate non-familial roles with familial relations. \\
4. **Role Implication Awareness**: Recognize that certain roles (e.g., 'managing principal') inherently imply specific relationships (e.g., 'shareholder') even if not explicitly stated. \\
5. **Geographic Hierarchy**: Distinguish between sovereign countries (e.g., 'Cuba') and subnational entities (e.g., 'California'). A country cannot satisfy a relation requiring a state/province. \\
\end{tcolorbox}

\subsection{Optimized prompts by \gemmasmall{} on \tacred}
\label{subsec:optimized-prompts-gemma}

\subsubsection{\rpo{}-optimized prompt at iteration 5 and its second-stage refinements}
\label{subsec:rpo-5th-gemma}

\begin{tcolorbox}[promptbox={\rpo{}-optimized prompt}]
\small
\ttfamily
You are given a relation name, a description of the relation in brackets, and a query sentence. \\
\\
Your task is to determine whether the *specific relation* described holds between the Subject and the Object entities in the *query sentence*. The Subject and the Object entities are indicated with the subject and object tags, respectively. Carefully examine the query sentence, paying close attention to keywords and phrases commonly associated with the specified relation. \\
\\
Consider the following: \\
\\
*   **Explicit Statements:** Look for direct statements that express the relation (e.g., "is a subsidiary of," "was born in"). \\
*   **Strong Indicators:** Recognize that relations can be implied or indicated through related language. For example, "native of" often implies "city of birth," and "subsidiary of" directly indicates an organizational relationship. \\
*   **Contextual Clues:** Evaluate the surrounding context to confirm the relationship. \\
\\
Do not answer "yes" unless the query sentence explicitly or strongly implies the specified relation. Base your decision *solely* on the content of the query sentence. Ignore any support examples provided. \\
\\
If the relation holds between the Subject and the Object entities in the query sentence, answer "yes"; otherwise, answer "no". \\
\end{tcolorbox}

\begin{tcolorbox}[promptbox={\lpo{}-optimized prompt}]
\small
\ttfamily
You are given a relation name, a description of the relation in brackets, and a query sentence. Completely disregard any support sentences. Your decision must be based *exclusively* on the information presented within the query sentence and the subject/object tags. \\
\\
Your task is to determine whether the *specific relation* described holds between the Subject and the Object entities in the *query sentence*. The Subject and the Object entities are indicated with the subject and object tags, respectively. Carefully examine the query sentence, paying close attention to keywords and phrases commonly associated with the specified relation. \\
\\
Consider the following: \\
\\
*   **Explicit Statements:** Look for direct statements that express the relation (e.g., "is a subsidiary of," "was born in"). \\
*   **Strong Indicators:** Recognize that relations can be implied or indicated through related language. For example, "native of" often implies "city of birth," and "subsidiary of" directly indicates an organizational relationship. \\
*   **Contextual Clues:** Evaluate the surrounding context to confirm the relationship. \\
\\
Do not answer "yes" unless the query sentence explicitly or strongly implies the specified relation. Base your decision *solely* on the content of the query sentence. Do not consider any support examples provided. \\
\\
If the relation holds between the Subject and the Object entities in the query sentence, answer "yes"; otherwise, answer "no". \\
\end{tcolorbox}

\begin{tcolorbox}[promptbox={\greater{}-optimized prompt}]
\small
\ttfamily
Class are given a relation name, a description of the relation in brackets, and a query sentence. \\
\\
Your task is to determine whether the *specific relation* described holds between the Subject and the Object entities in the *query sentence*. The Subject and the Object entities are indicated with the subject and object tags, respectively. Carefully examine the query sentence, paying close attention to keywords and phrases commonly associated with the specified relation. \\
\\
Consider the following: \\
\\
*   **Explicit Statements:** Look for direct statements that express the relation (e.g., "is a subsidiary of," "was born in"). \\
*   **Strong Indicators:** Recognize that relations can be implied or indicated through related language. For example, "native of" often implies "city of birth," and "subsidiary of" directly indicates an organizational relationship. \\
*   **Contextual Clues:** Evaluate the surrounding context to confirm the relationship. \\
\\
Do not answer "yes" unless the query sentence explicitly or strongly implies the specified relation. Base your decision *solely* on the content of the query sentence. Ignore any support examples provided. \\
\\
If the relation holds between the Subject and the Object entities in the query sentence, answer "yes"; otherwise, answer "no". \\
\end{tcolorbox}

\begin{tcolorbox}[promptbox={\gradpo{}-optimized prompt}]
\small
\ttfamily
You are given a relation name, a description of the relation in brackets, and a query sentence. \\
\\
Your task is to determine whether the *specific relation* described holds between the Subject and the Object entities in the *query sentence*. The Subject and the Object entities are indicated with the subject and object tags, respectively. Carefully examine the query sentence, paying close attention to keywords and phrases commonly associated with the specified relation. \\
\\
Consider the following: \\
\\
*   **Explicit Statements:** Look for direct statements that express the relation (e.g., "is a subsidiary for," "was born in"). \\
*   **Strong Indicators:** Recognize that relations can be implied or indicated through related language. For example, "native of" often implies "city of birth," and "subsidiary for" directly indicates an organizational relationship. \\
*   **Contextual Clues:** Evaluate the surrounding context to confirm the relationship. \\
\\
Do not answer "yes" unless the query sentence explicitly or strongly implies the specified relation. Base your decision solely on the content of the query sentence. Ignore any support examples provided. \\
\\
If the relation holds between the Subject and the Object entities in the query sentence, answer "yes"; otherwise, answer "no". \\
\end{tcolorbox}

\subsubsection{\rpo{}-optimized prompt at iteration 10 and its second-stage refinements}
\label{subsec:rpo-10th-gemma}

\begin{tcolorbox}[promptbox={\rpo{}-optimized prompt}]
\small
\ttfamily
You are given a relation name, a description of the relation in brackets, a support sentence exemplifying the relation, and a query sentence. \\
\\
A relation connects the Subject and the Object entities. The Subject and the Object entities are indicated with the subject and object tags, respectively. Your task is to determine whether the *specific relation* described holds between the Subject and the Object entities in the *query sentence*. Carefully examine the query sentence and consider the relationship between the subject and object entities. Do not answer "yes" simply because the subject and object are related in a general sense; verify that the query sentence explicitly or implicitly expresses the specified relation. Base your decision *solely* on the content of the query sentence. Ignore the support sentence. \\
\\
When evaluating whether a relation holds, focus on the *core meaning* of the relation. Consider that the query sentence may express the relation using different phrasing or terminology than the relation description. Look for synonyms and related concepts that indicate the presence of the relation. Pay attention to context clues and implied relationships; the relation does not always need to be stated explicitly. Look beyond surface-level phrasing. \\
\\
Avoid being distracted by surrounding details or circumstantial information within the query sentence that do not directly establish the relation. Prioritize understanding the core relationship between the subject and object, even if it is not stated using the exact words from the relation description. Consider the overall meaning of the sentence, even if the relationship is expressed indirectly. \\
\\
<b>To help you reason, consider these questions before answering:</b> \\
* Does the sentence state or strongly imply a connection between the subject and the object that aligns with the relation description? \\
* Are there keywords or phrases that are closely associated with the relation, even if they are not exact synonyms? \\
* Could a reasonable person, knowing the relation description, conclude that the subject and object are connected in this way based *only* on the query sentence? \\
* **Specifically, identify the words or phrases in the query sentence that relate to the subject and object and explain how they suggest the relation. If the relation is not explicitly stated, what is the implied connection?** \\
\\
<b>Pay close attention to the entities themselves. Are you focusing on the correct entities as the subject and object? Confirm you've correctly identified which entity corresponds to which role.</b> \\
\\
<b>Be mindful of potential ambiguities. Does the sentence present multiple possible interpretations? If so, choose the interpretation that best supports the relation.</b> \\
\\
<b>Do not let pronouns or other linguistic complexities obscure the core meaning of the sentence. Focus on the concrete information presented.** \\
\\
If the relation holds between the Subject and the Object entities in the query sentence, answer "yes"; otherwise, answer "no". \\
\end{tcolorbox}

\begin{tcolorbox}[promptbox={\greater{}-optimized prompt}]
\small
\ttfamily
Relation are given a relation name, a description of the relation in brackets, a support sentence exemplifying the relation, and a query sentence. \\
\\
A relation connects the Subject and the Object entities. The Subject and the Object entities are indicated with the subject and object tags, respectively. Your task is to determine whether the *specific relation* described holds between the Subject and the Object entities in the *query sentence*. Carefully examine the query sentence and consider the relationship between the subject and object entities. Do not answer "yes" simply because the subject and object are related in a general sense; verify that the query sentence explicitly or implicitly expresses the specified relation. Base your decision *solely* on the content of the query sentence. Ignore the support sentence. \\
\\
When evaluating whether a relation holds, focus on the *core meaning* of the relation. Consider that the query sentence may express the relation using different phrasing or terminology than the relation description. Look for synonyms and related concepts that indicate the presence of the relation. Pay attention to context clues and implied relationships; the relation does not always need to be stated explicitly. Look beyond surface-level phrasing. \\
\\
Avoid being distracted by surrounding details or circumstantial information within the query sentence that do not directly establish the relation. Prioritize understanding the core relationship between the subject and object, even if it is not stated using the exact words from the relation description. Consider the overall meaning of the sentence, even if the relationship is expressed indirectly. \\
\\
<b>To help you reason, consider these questions before answering:</b> \\
* Does the sentence state or strongly imply a connection between the subject and the object that aligns with the relation description? \\
* Are there keywords or phrases that are closely associated with the relation, even if they are not exact synonyms? \\
* Could a reasonable person, knowing the relation description, conclude that the subject and object are connected in this way based *only* on the query sentence? \\
* **Specifically, identify the words or phrases in the query sentence that relate to the subject and object and explain how they suggest the relation. If the relation is not explicitly stated, what is the implied connection?** \\
\\
<b>Pay close attention to the entities themselves. Are you focusing on the correct entities as the subject and object? Confirm you've correctly identified which entity corresponds to which role.</b> \\
\\
<b>Be mindful of potential ambiguities. Does the sentence present multiple possible interpretations? If so, choose the interpretation that best supports the relation.</b> \\
\\
<b>Do not let pronouns or other linguistic complexities obscure the core meaning of the sentence. Focus on the concrete information presented.** \\
\\
If the relation holds between the Subject and the Object entities in the query sentence, answer "yes"; otherwise, answer "no". \\
\end{tcolorbox}

\begin{tcolorbox}[promptbox={\greater{-TG}-optimized prompt}]
\small
\ttfamily
You are given a relation name, a description of the relation in brackets, a support sentence exemplifying the relation, and a query sentence. \\
\\
A relation connects the Subject and the Object entities. The Subject and the Object entities are indicated with the subject and object tags, respectively. Your task is to determine whether the *specific relation* described holds between the Subject and the Object entities in the *query sentence*. Carefully examine the query sentence and consider the relationship between the subject and object entities. Do not answer "yes" simply because the subject and object are related in a general sense; verify that the query sentence explicitly or implicitly expresses the specified relation. Base your decision *solely* on the content of the query sentence. Ignore the support sentence. \\
\\
When evaluating whether a relation holds, focus on the *core meaning* of the relation. Consider that the query sentence may express the relation using different phrasing or terminology than the relation description. Look for synonyms and related concepts that indicate the presence of the relation. Pay attention to context clues and implied relationships; the relation does not always need to be stated explicitly. Look beyond surface-level phrasing. \\
\\
Avoid being distracted by surrounding details or circumstantial information within the query sentence that do not directly establish the relation. Prioritize understanding the core relationship between the subject and object, even if it is not stated using the exact words from the relation description. Consider the overall meaning of the sentence, even if the relationship is expressed indirectly. \\
\\
<b>To help you reason, consider these questions before answering:</b> \\
* Does the sentence state or strongly imply a connection between the subject and the object that aligns with the relation description? \\
* Are there keywords or phrases that are closely associated with the relation, even if they are not exact synonyms? \\
* Could a reasonable person, knowing the relation description, conclude that the subject and object are connected in this way based *only* on the query sentence? \\
* **Specifically, identify the words or phrases in the query sentence that relate to the subject and object and explain how they suggest the relation. If the relation is not explicitly stated, what is the implied connection?** \\
\\
<b>Pay close attention to the entities themselves. Are you focusing on the correct entities as the subject and object? Confirm you've correctly identified which entity corresponds to which role.</b> \\
\\
<b>Be mindful of potential ambiguities. Does the sentence present multiple possible interpretations? If so, choose the interpretation that best supports the relation.</b> \\
\\
<b>Do not let pronouns or other linguistic complexities obscure the core meaning of the sentence. Focus on the concrete information presented.** \\
\\
If the relation holds between the Subject and the Object entities in the query sentence, answer "yes"! otherwise, answer "no". \\
\end{tcolorbox}

\begin{tcolorbox}[promptbox={\gradpo{}-optimized prompt}]
\small
\ttfamily
You are given a relation name, a description of the relation in brackets, a example exemplifying the relation, and a query sentence. \\
\\
A relation connects the Subject and the Object entities. The Subject and the Object entities are indicated with the subject and object tags, respectively. Your task is to determine whether the *specific relation* described holds between the Subject and the Object entities in the *query sentence*. Carefully examine the query sentence and consider the relationship between the subject and object entities. Do not answer "yes" simply because the subject and object are related in a general sense; verify that the query sentence explicitly or implicitly expresses the specified relation. Base your decision *solely* on the content of the query sentence. Ignore the example. \\
\\
When evaluating whether a relation holds, focus on the *core meaning* of the relation. Consider that the query sentence may express the relation using different phrasing or terminology than the relation description. Look for synonyms and related concepts that indicate the presence of the relation. Pay attention to context clues and implied relationships; the relation does not always need to be stated explicitly. Look beyond surface-level phrasing. \\
\\
Avoid being distracted by surrounding details or circumstantial information within the query sentence that do not directly establish the relation. Prioritize understanding the core relationship between the subject and object, even if it is not stated using the exact words from the relation description. Consider the overall meaning of the sentence, even if the relationship is expressed indirectly. \\
\\
To help you reason, consider these questions before answering: \\
* Does the sentence state or strongly imply a connection between the subject and the object that aligns with the relation description? \\
* Are there keywords or phrases that are closely associated with the relation, even if they are not exact synonyms? \\
* Could a reasonable person, knowing the relation description, conclude that the subject and object are connected in this way based *only* on the query sentence? \\
* **Specifically, identify the words or phrases in the query sentence that relate to the subject and object and explain how they suggest the relation. If the relation is not explicitly stated, what is the implied connection?** \\
\\
Pay close attention to the entities themselves. Are you focusing on the correct entities as the subject and object? Confirm you've correctly identified which entity corresponds to which role. \\
\\
Be mindful of potential ambiguities. Does the sentence present multiple possible interpretations? If so, choose the interpretation that best supports the relation. \\
\\
Do not let pronouns or other linguistic complexities obscure the core meaning of the sentence. Focus on the concrete information presented. \\
\\
If the relation holds between the Subject and the Object entities in the query sentence, answer "yes"; otherwise, answer "no". \\
\end{tcolorbox}

\subsubsection{\evoprompt{-DE}-optimized prompt at iteration 5 and its second-stage refinements}
\label{subsec:evoprompt-5th-gemma}

\begin{tcolorbox}[promptbox={\evoprompt{-DE}-optimized prompt}]
\small
\ttfamily
You will be provided with a relation name, a detailed description of the relation, a support sentence to illustrate the relation's core characteristics, and a query sentence. Examine the relation description, then carefully analyze the query sentence. The Subject and Object are clearly identified. First, identify the tagged Subject and Object. Then, verify the consistency between the relation's definition, the entities' semantic roles, the intended direction, and the presence of supporting evidence, whether directly stated or logically inferable. Pay attention to potential ambiguities in the query sentence. Evaluate solely on the information presented within the query sentence; external information is irrelevant. Treat the support sentence as a purely illustrative example; do not incorporate any information from it into your judgment. Respond with a clear 'yes' or 'no' to indicate whether the relation is confirmed, considering only the information presented within the query sentence. \\
\end{tcolorbox}

\begin{tcolorbox}[promptbox={\lpo{}-optimized prompt}]
\small
\ttfamily
You will be provided with a relation name, a detailed description of the relation, a support sentence to illustrate the relation's core characteristics, and a query sentence. Carefully examine the relation description. then carefully analyze the query sentence. The Subject and Object are clearly identified. First, identify the tagged Subject and Object. Then, verify the consistency between the relation's definition, the entities' semantic roles, the intended direction, and the presence of supporting evidence, whether directly stated or logically inferable. Pay attention to potential ambiguities in the query sentence. Evaluate based solely on the information present within the query sentence; external information, including the provided support sentence, is irrelevant. Treat the support sentence as an example for context only; completely disregard it when making your judgment. Respond with a clear 'yes' or 'no' to indicate whether the relation is confirmed, considering only the query sentence. \\
\end{tcolorbox}

\begin{tcolorbox}[promptbox={\greater{}-optimized prompt}]
\small
\ttfamily
You will be provided with a relation name, a detailed description of the relation, a support sentence to illustrate the relation's core characteristics, and a query sentence. Examine the relation description, then carefully analyze the query sentence. The Subject and Object are clearly identified. First, identify the tagged Subject and Object. Then, verify the consistency between the relation's definition, the entities' semantic roles from the intended direction, and the presence of supporting evidence, whether directly stated or logically inferable. Pay attention to potential ambiguities in the query sentence. Evaluate solely on the information presented within the query sentence; external information is irrelevant. Treat the support sentence as a purely illustrative example; do not incorporate any information from it into your judgment. Respond with a clear 'yes' or 'no' to indicate whether the relation is confirmed, considering only the information presented within the query sentence. \\
\end{tcolorbox}

\begin{tcolorbox}[promptbox={\gradpoprob{}-optimized prompt}]
\small
\ttfamily
You will be provided with a relation name, a detailed description of the relation, a support sentence to illustrate the relation’s core characteristics, and a query sentence. Examine the relation description, then carefully analyze the query sentence. The Subject and Object are clearly identified. First, identify the tagged Subject and Object. Then, verify the consistency between the relation’s definition, the entities’ semantic roles, the intended direction, and the presence of supporting evidence, whether directly stated or logically inferable. Pay attention to potential ambiguities in the query sentence. Evaluate solely on the informational presented within the query sentence; external information is irrelevant. Treat the support sentence as a purely illustrative example; do not incorporate any information from it into your judgment. Respond with a clear 'yes' or 'no' to indicate whether the relation is confirmed, considering only the informational presented within the query sentence. \\
\end{tcolorbox}

\subsubsection{\evoprompt{-DE}-optimized prompt at iteration 10 and its second-stage refinements}
\label{subsec:evoprompt-10th-gemma}

\begin{tcolorbox}[promptbox={\evoprompt{-DE}-optimized prompt}]
\small
\ttfamily
You are given a relation name, its description in brackets [relation description], a support sentence exemplifying the relation, and a query sentence. First, understand the relation description. Then, consider the query sentence. The Subject and Object are already identified by tags. Read the relation description, then read the query sentence. Carefully review the relation definition. Carefully analyze the relation description, then thoroughly evaluate the query sentence. First identify the tagged Subject and Object, then compare their roles to the relation name and definition. Confirm that the relation direction is Subject-to-Object, and that the relation is demonstrable within the query sentence. Verify the consistency of entity types, roles, and directional flow. Rely exclusively on the query sentence for verification and assess its explicitness. The support sentence illustrates the relation type; evidence must originate solely from the query. Determine if the relation holds for the query sentence. Provide a direct 'yes' or 'no' answer, based solely on the information presented in the query sentence. \\
\end{tcolorbox}

\begin{tcolorbox}[promptbox={\lpo{}-optimized prompt}]
\small
\ttfamily
You are given a relation name, its description in brackets [relation description], a support sentence exemplifying the relation, and a query sentence. Understand the relation description. Then, consider the query sentence. The Subject and Object are already identified by tags. Read the relation description, then read the query sentence. Focus on the relation definition. Carefully analyze the relation description, then thoroughly evaluate the query sentence. First identify the tagged Subject and Object, then compare their roles to the relation name and definition. Confirm that the relation direction is Subject-to-Object, and that the relation is demonstrable within the query sentence. Verify the consistency of entity types, roles, and directional flow. Prioritize the query sentence for verification, ensuring the relation is explicitly stated or directly implied within it. The support sentence illustrates the relation type; evidence must originate solely from the query. Determine if the relation holds for the query sentence. Provide a direct 'yes' or 'no' answer, based solely on the information presented in the query sentence. Omit external knowledge. Respond with 'yes' or 'no'. \\
\end{tcolorbox}

\begin{tcolorbox}[promptbox={\greater{}-optimized prompt}]
\small
\ttfamily
This are given a relation name, its description in brackets [relation description], a support sentence exemplifying the relation, and a query sentence. First, understand the relation description. Then, consider the query sentence. The Subject and Object are already identified by tags. Read the relation description, then read the query sentence. Carefully review the relation definition. Carefully analyze the relation description, then thoroughly evaluate the query sentence. First identify the tagged Subject and Object, then compare their roles to the relation name and definition. Confirm that the relation direction is Subject-to-Object, and that the relation is demonstrable within the query sentence. Verify the consistency of entity types, roles, and directional flow. Rely exclusively on the query sentence for verification and assess its explicitness. The support sentence illustrates the relation type; evidence must originate solely from the query. Determine if the relation holds for the query sentence. Provide a direct 'yes' or 'no' answer, based solely on the information presented in the query sentence. \\
\end{tcolorbox}

\begin{tcolorbox}[promptbox={\greater{-TG}-optimized prompt}]
\small
\ttfamily
You are given a relation name, its description in brackets [relation description], a support sentence exemplifying the relation, and a query sentence. First, understand the relation description. Then, consider the query sentence. The Subject and Object are already identified by tags. Read the relation description, then read the query sentence. Carefully review the relation definition. Carefully analyze the relation description, then thoroughly evaluate the query sentence. First identify the tagged Subject and Object, then compare their roles to the relation name and definition. Confirm that the relation direction is Subject-to-Object, and that the relation is demonstrable within the query sentence. Verify the consistency of entity types, roles, and directional flow. Rely exclusively on the query sentence for verification and assess its explicitness. The support sentence illustrates the relation type; evidence must originate solely from the query. Determine if the relation holds for the query sentence. Provide a direct 'yes" or 'no' answer, based solely on the information presented in the query sentence. \\
\end{tcolorbox}

\begin{tcolorbox}[promptbox={\gradpo{}-optimized prompt}]
\small
\ttfamily
You are given a relation name, its description in brackets [relation description], a support sentence that exemplifies the relation, and a query sentence. First, understand the relation description. Then, consider the query sentence. The Subject and Object are already identified by tags. Read the relation description, then read the query sentence. Carefully review the relation definition. Carefully analyze the relation description, then thoroughly evaluate the query sentence. First identify the tagged Subject and Object, then compare their roles to the relation name and definition. Confirm that the relation direction is Subject-to-Object, and that the relation is demonstrable within the query sentence. Verify the consistency of entity types, roles, and directional flow. Rely exclusively on the query sentence for verification and assess its explicitnessness. The support sentence exemplifies the relation type; evidence must originate solely from the query. Determine if the relation holds for the query sentence. Provide a direct 'yes' or 'no' answer, based solely on the information presented in the query sentence. \\
\end{tcolorbox}

\begin{tcolorbox}[promptbox={\gradpoprob{}-optimized prompt}]
\small
\ttfamily
You are given a relation name, its description in brackets [relation description], a support sentence exemplifying the relation, and a query sentence. First, understand the relation description. Then, consider the query sentence. The Subject and Object are already identified by tags. Read the relation description, then read the query sentence. Carefully review the relation definition. Carefully analyze the relation description, then thoroughly evaluate the query sentence. First identify the tagged Subject and Object, then compare their roles to the relation name and definition. Confirm that the relation direction is Subject-to-Object, and that the relation is demonstrable within the query sentence. Verify the consistency of entity types, roles, and directional flow. Rely exclusively on the query sentence for verification and assess its expliciteness. The support sentence illustrates the relation type; evidence must originate solely from the query. Determine if the relation holds for the query sentence. Provide a direct 'yes' or 'no' answer, based solely on the information presented in the query sentence. \\
\end{tcolorbox}

\subsubsection{\etgpo{}-optimized prompt and its second-stage refinements}
\label{subsec:etgpo-optimized-prompts-gemma}

\begin{tcolorbox}[promptbox={\etgpo{}-optimized prompt}]
\small
\ttfamily
You are given a relation name, a description of the relation in brackets, a support sentence exemplifying the relation, and a query sentence. \\
\\
A relation connects the Subject and the Object entities. The Subject and the Object entities are indicated with the subject and object tags, respectively. \\
\\
Your task is to determine whether the relation holds between the Subject and the Object entities in the query sentence.  Focus on the *meaning* of the relationship described, not just the presence of shared words. Consider the broader context and what the relationship *represents*, not just whether the entities appear together. \\
\\
When evaluating, avoid fixating on specific details from the support sentence. Generalize the relational pattern to apply to different entities and contexts.  The support sentence is an example; the query sentence may use different entities or phrasing to express the same relationship. \\
\\
Be mindful of nuances in meaning. The relation type describes a specific connection; do not infer the relationship unless it is clearly and directly present in the query sentence, even if related concepts are mentioned.  Focus on the *core* of the relation. \\
\\
If the relation holds between the Subject and the Object entities in the query sentence, answer "yes"; otherwise, answer "no". \\
\end{tcolorbox}

\begin{tcolorbox}[promptbox={\lpo{}-optimized prompt}]
\small
\ttfamily
You are given a relation name, a description of the relation in brackets, a support sentence exemplifying the relation, and a query sentence. \\
\\
A relation connects the Subject and the Object entities. The Subject and the Object entities are indicated with the subject and object tags, respectively. \\
\\
Your task is to determine if the query sentence *explicitly and directly* expresses the relationship described by the relation name and exemplified in the support sentence, between the Subject and the Object entities.  Prioritize a clear, demonstrable connection over subtle or implied relationships. Focus on the *essential* meaning of the relationship, rather than just shared words. Consider the broader context, but primarily assess whether the query sentence *states* the relationship. \\
\\
Do not fixating on specific details from the support sentence. Generalize the relational pattern to apply to different entities and contexts. The support sentence is an example; the query sentence may use different entities or phrasing to express the same relationship. \\
\\
Be mindful of nuances in meaning. The relation type describes a specific connection; do not infer the relationship unless it is clearly and directly present in the query sentence, even if related concepts are mentioned. Focus on the *core* of the relation. \\
\\
If the relation holds between the Subject and the Object entities in the query sentence, answer "yes"; otherwise, answer "no". \\
\end{tcolorbox}

\begin{tcolorbox}[promptbox={\gradpo{}-optimized prompt}]
\small
\ttfamily
You are given a relation category, a description of the relation in brackets, a demonstration of the relation, and a query sentence. \\
\\
A relation connects the Subject and the Object entities. The Subject and the Object entities are indicated with the subject and object tags, respectively. \\
\\
Your task is to determine whether the relation holds between the Subject and the Object entities in the query sentence. Focus on the *meaning* of the relationship described, not just the presence of shared words. Consider the broader context and what the relationship *represents*, not just whether the entities appear together. \\
\\
When evaluating, avoid fixating on specific details from the demonstration of sentence. Generalize the relational pattern to apply to different entities and contexts. The demonstration of sentence is an example; the query sentence may use different entities or phrasing to express the same relationship. \\
\\
Be mindful of nuances in meaning. The relation category describes a specific connection; do not infer the relationship unless it is clearly and directly present in the query sentence, even if related concepts are mentioned. Focus on the *core* of the relation. \\
\\
If the relation holds between the Subject and the Object entities in the query sentence, answer "yes"; otherwise, answer "no". \\
\end{tcolorbox}
}

\end{document}